\documentclass{article} 
\usepackage{iclr2027_conference,times}

\usepackage{amsmath,amsfonts,bm}

\def\eqref#1{equation~\ref{#1}}

\def\1{\bm{1}}

\DeclareMathAlphabet{\mathsfit}{\encodingdefault}{\sfdefault}{m}{sl}
\SetMathAlphabet{\mathsfit}{bold}{\encodingdefault}{\sfdefault}{bx}{n}

\usepackage{hyperref}
\usepackage{url}
\usepackage{subcaption}
\usepackage{booktabs}
\usepackage[table]{xcolor}

\usepackage{tcolorbox}
\tcbuselibrary{breakable}

\newtcolorbox{hypothesisbox}{
  breakable,
  colback=black!2,
  colframe=black!55,
  boxrule=0pt,
  leftrule=1.5pt,
  sharp corners,
  left=9pt,
  right=9pt,
  top=7pt,
  bottom=7pt,
  before skip=10pt,
  after skip=10pt
}

\title{Competing Learning in Hybrid Architectures}

\iclrfinalcopy
\author{Xingyu Shen$^*$, Yingfa Chen$^*$, Ziqing Qiao, Chaojun Xiao, Xu Han, \& Zhiyuan Liu \\
    Department of Science and Technology, Tsinghua University, Beijing, China\\
    \texttt{xingyu-c21@mails.tsinghua.edu.cn}, \texttt{chenyingfa1999@gmail.com}
}

\usepackage{xspace}

\newcommand{\OURS}{BLD\xspace}
\newcommand{\KEY}{Competing Learning\xspace}
\newcommand{\metric}{retrieval ability\xspace}

\begin{document}

\maketitle

\begin{abstract}


Many modern language models have adopted hybrid architectures that incorporate both linear-time recurrence and softmax attention. Modern recurrence layers provide a highly expressive mixing mechanism while ensuring high throughput by compressing contextual information into a fixed-size state, but underperform softmax attention in contextual information retrieval. In this paper, we reveal a \textit{Competing Learning} phenomenon where different kinds of mixers compete to learn the same function, and when any layer learns a function, it suppresses the corresponding learning signal for other layers. Layer attribution analysis reveals that under typical data distributions, modern recurrence learns retrieval functions faster than softmax attention, causing attention to fail to learn retrieval in most cases. Stopping weight updates in recurrence in the middle of training helps attention learn retrieval. Based on these results, we propose a staged training procedure for pre-training hybrid architectures on natural language. Empirical results on models up to 1B parameters show that our method results in substantial improvement in language modeling loss, common-sense reasoning, as well as a wide range of retrieval-related tasks across different context lengths.~\footnote{The code is released at \url{https://github.com/thunlp/competing-learning}.}


\end{abstract}

\section{Introduction}

Recently, many open-weight large language models~(LLMs) have adopted \textit{hybrid architectures} that combine two distinct types of token mixers: linear recurrence~(e.g., state space models~(SSMs) and linear attention) and softmax attention~\citep{vaswani2017attention}. Such models have demonstrated strong language modeling capabilities while offering improved training and inference efficiency, scaling to models with up to trillions of parameters. Representative examples include Kimi K3~\citep{team2026kimi}, GLM-5.3-Flash~\citep{glm-5_3-flash}, Qwen3.8-Flash-Next~\citep{qwen3.8flashnext}, and Nemotron-3-Ultra~\citep{nemotron-3-ultra}.


Prior work has shown that, in well-trained LLMs, different token mixers---and/or different heads within the same mixer---can specialize in distinct mixing patterns. For instance, \citet{xiao2025duoattention} show that a subset of attention heads in Transformer models~\citep{vaswani2017attention}, referred to as \textit{retrieval heads}, learn long-distance mixing patterns, whereas other heads primarily capture short-distance interactions. Similarly, \citet{bick2025understanding} show that different heads within a linear recurrence layer exhibit retrieval behaviors over different distances. In hybrid architectures, \citet{chen2026hybrid} further show that assigning long-distance retrieval to softmax attention and short-distance mixing to recurrent layers is beneficial across a wide range of downstream tasks. Together, these findings motivate a division of labor in hybrid models, where softmax attention specializes in long-distance mixing while recurrent layers specialize in short-distance mixing.



In this work, we identify \textit{\KEY{}}, a phenomenon in which layers that acquire retrieval capabilities earlier inhibit the development of these capabilities in other layers. In the hybrid models we study, recurrent layers tend to gain this early advantage, taking over retrieval-related functions that softmax attention layers are intended to specialize in. Consequently, training hybrid models using standard practices results in softmax attention's retrieval learning being limited by recurrent layers, leading to suboptimal division of labor.

\begingroup
\renewcommand{\thefootnote}{\fnsymbol{footnote}}
\footnotetext[1]{Equal contributions.}
\endgroup

Concretely, we provide an analysis based on layer-wise contribution attribution on synthetic retrieval tasks to identify the speed of retrieval learning of different mixers in a hybrid model. The analysis provides evidence to illustrate how the learning of retrieval functions in softmax attention is inhibited by the retrieval learning of recurrent layers. Furthermore, manually stopping recurrent layers from learning retrieval at different points in the training process reveals that softmax attention starts learning retrieval as soon as recurrent layers' learning is stopped. We also discover that under typical training distributions, with more short-distance patterns than long-distance ones, recurrence learns retrieval faster than softmax attention, resulting in suboptimal division of labor.

To address this issue, we propose Better Layer Division~(\OURS{}), a training pipeline that introduces a preliminary training stage before standard pretraining to give full-attention layers an early advantage in learning retrieval. This intervention is designed to steer \KEY{} in favor of full-attention layers, encouraging them to assume the long-range retrieval role during subsequent pretraining. We evaluate \OURS{} on widely used benchmarks (e.g. Common-sense reasoning (CSR), recall-intensive tasks, LongPPL~\citep{fang2025wrong}.) using hybrid models based on Kimi Delta Attention (KDA)\footnote{Widely-used RNN architecture adopted by frontier models (e.g. Kimi K3, GLM-5.3-Flash).}~\citep{team2025kimi} at multiple scales, and observe improvements in benchmarks including CSR, recall-intensive tasks, RULER, LongPPL and training loss over vanilla ones.

Our main contributions are as follows:
\begin{itemize}
    \item We investigate the training dynamics of layer-wise functional specialization in hybrid models, focusing on the allocation of retrieval-related functions between recurrent and full-attention layers.

    \item We identify and characterize \KEY{} through empirical analyses, revealing how early differences in retrieval learning influence subsequent layer specialization.

    \item We propose \OURS{}, an LLM training pipeline that guides functional specialization by freezing the parameters of selected components during the early stage of pretraining, and verify its effectiveness on hybrid models at multiple scales and with different recurrent architectures. Empirical results show that \OURS{} produces pretrained LLMs with performance gains on a wide range of tasks.
\end{itemize}

\section{Related Work}

\paragraph{Hybrid Architectures}

Recent LLMs increasingly adopt hybrid architectures that combine softmax attention with linear attention, including GLM-5.3-Flash~\citep{glm-5_3-flash}, Kimi K3~\citep{team2026kimi}, Qwen3.8-Flash-Next~\citep{qwen3.8flashnext}, and Nemotron-3-Ultra~\citep{nemotron-3-ultra}. Such designs combine the retrieval strength of softmax attention with the computational efficiency of linear attention. Prior work has shown that linear or recurrent models can underperform softmax attention on precise in-context retrieval~\citep{arora2024simple,bick2025understanding,jelassi2024repeat}. Several works improve the memory or retrieval capabilities of linear attention through architectural or state-space enhancements~\citep{yang2023gated,yang2025gated,qin2024hgrn2,shen2026statex}. In contrast, our work does not enhance the retrieval capacity of linear attention itself, but improves the pre-training of hybrid models without architectural modifications. Hybrid architectures instead retain softmax-attention layers for long-range retrieval while using efficient linear attention for most computation~\citep{team2025kimi}. Our work focuses on better realizing this architectural complementarity during pre-training. Most of our experiments are conducted on Kimi-Linear-style models based on Kimi Delta Attention (KDA)~\citep{team2025kimi}.

\paragraph{Labor Division in Non-Hybrid Models}

Prior work has shown that only a small subset of attention heads develops strong long-context retrieval behavior~\citep{wu2025retrieval}. DuoAttention exploits this functional specialization by distinguishing retrieval heads from streaming heads for efficient inference~\citep{xiao2025duoattention}. Other works leverage already-formed retrieval heads to construct post-training signals for improving long-context performance and contextual faithfulness~\citep{huang2025improving,ma2026interpretability}. Recent studies further analyze how retrieval heads are learned or evolve during context extension and long-context continual pre-training~\citep{zhao2024understanding,liang2026revealing}. These works mainly study or exploit specialization among homogeneous softmax-attention heads, rather than the allocation of retrieval capability across heterogeneous attention components. In hybrid models, however, such allocation is critical: if all retrieval heads emerge in linear-attention components, the intended architectural advantage of combining efficient linear attention with the stronger retrieval capability of softmax attention would be largely undermined.

\paragraph{Labor Division in Hybrid Models}

Prior work has observed functional specialization between attention and recurrent components in hybrid models, with attention playing a particularly important role in exact retrieval and recall~\citep{bick2025understanding,afendulev2026attention,wang2025systematic}. These works mainly characterize specialization after it has emerged, while other studies exploit such specialization to construct hybrid models from pretrained Transformers~\citep{chen2026hybrid,bick2026retrieval,li2026distilling}. In contrast, we study how this division of labor forms during hybrid-model pre-training and how it can be actively optimized. In the work most closely related to ours, \citet{qiao2026rethinking} show that efficient-attention modules can weaken the optimization pressure for full-attention layers and delay retrieval-head formation. Our work further identifies the underlying competition in retrieval-specific learning signals: linear-attention and softmax-attention layers both gradually acquire retrieval capabilities, and layers that learn retrieval earlier reduce the retrieval-specific prediction error, thereby weakening the learning signal available to other layers. We refer to this as the \KEY{} phenomenon and use it to guide pre-training toward stronger retrieval specialization in softmax-attention layers.

\section{\KEY{} in Hybrid Attention Models}
\label{sec:division-of-labor}


Central to our paper is the \KEY~phenomenon, which is stated as follows:




\begin{hypothesisbox}
\noindent\textbf{\KEY{} Phenomenon}\par\smallskip
\noindent During hybrid-model pre-training, both recurrent and
softmax-attention layers can acquire in-context retrieval
capabilities. If some layers learn retrieval earlier, they can reduce the retrieval-specific prediction error, weakning the training signal for other layers, slowing their
acquisition of the same capability.
\end{hypothesisbox}

To show the causes and effects of this phenomenon, we conduct a series of layer attribution and gradient intervention experiments to monitor the retrieval learning of each layer in a hybrid model. In the following sections, we first propose a synthetic retrieval task~(Section~\ref{sec:retrieval-task}). Then, we propose a novel metric for determining how much a layer contributes to the model's retrieval abilities~(Section~\ref{sec:layer-contribution}). Finally, we train a series of hybrid models on this task and inspect the contributions of each layer to monitor their retrieval learning over time~(Section~\ref{sec:analysis-results}).

Concretely, we conduct the following three sets of experiments:
\begin{itemize}
    \item Section~\ref{sec:freeze_leading}: We provide the experimental evidence for the \KEY{} phenomenon with weight freezing.
    \item Section~\ref{sec:grad_change}: We provide direct gradient evidence for the \KEY{} phenomenon through weight freezing and unfreezing.
    \item Section~\ref{sec:fraction_speed}: We show how the distribution of short and long-distance patterns chooses the winner of \KEY{} phenomenon.
\end{itemize}
Altogether, we demonstrate the existence of \KEY{} in hybrid models and that KDA layers learn modeling short-distance patterns faster than attention layers. Since short-distance patterns are more frequent than long-distance patterns in natural language pretraining~\citep{doi:10.1073/pnas.1502134112}, this leads to a suboptimal labor division in direct pretraining of hybrid models.

\begin{figure}[!t]
    \centering
    \includegraphics[width=1\linewidth]{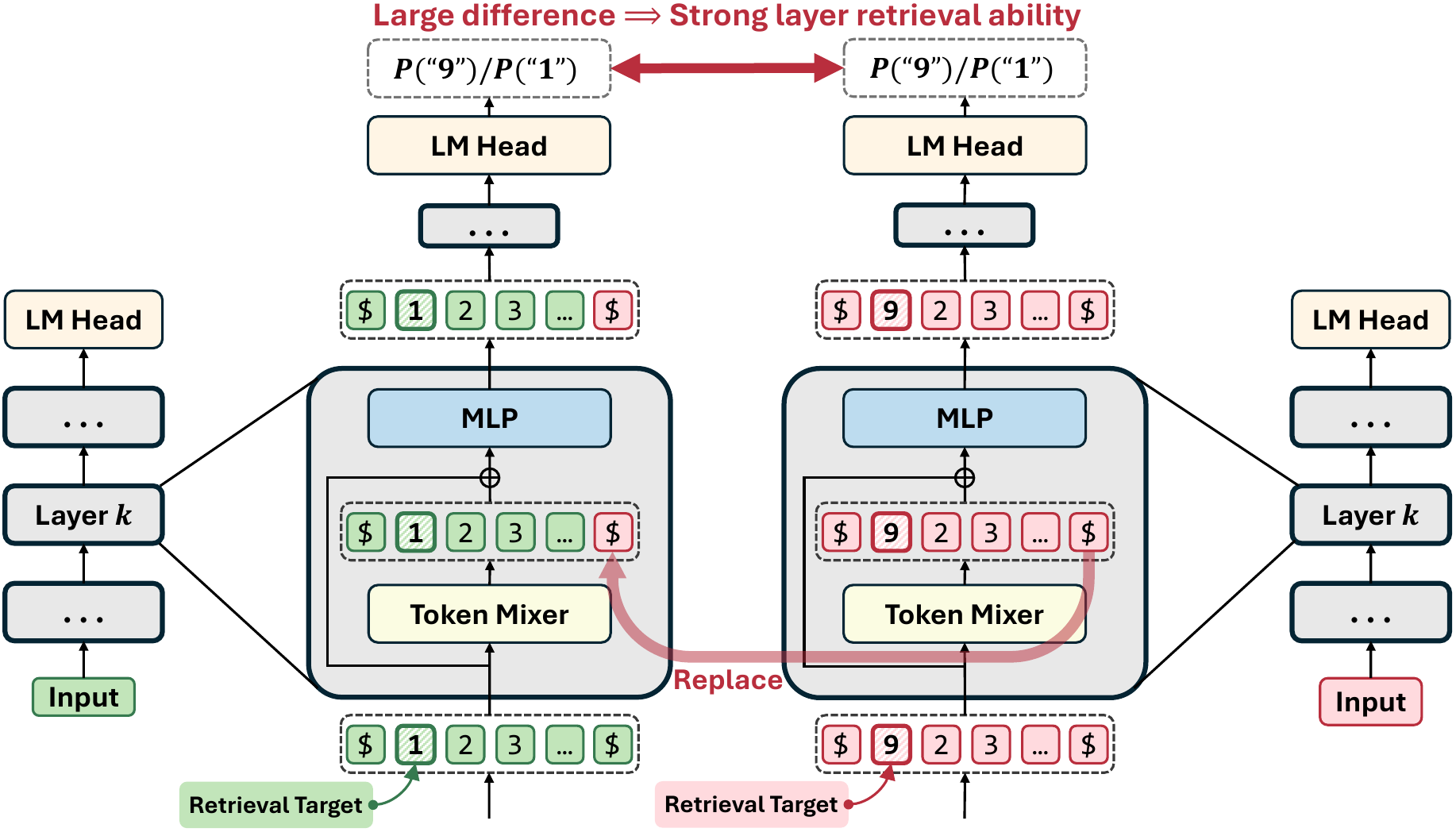}
    \caption{Illustration of our layer retrieval ability estimation method~(described in Section~\ref{sec:layer-contribution}). As an example, the sequence to copy is ``123'', and the layer under inspection is layer $k$. The difference in probability change tells us the contribution of layer $k$ in retrieving the target token ``9''.}
    \label{fig:layer_contribution}
\end{figure}

\subsection{Spaced Retrieval}
\label{sec:retrieval-task}

In order to evaluate retrieval ability without interference from complicated pattern in natural language, we craft a synthetic task called ``Spaced Retrieval'' in the following format:

\[
\begin{array}{r@{\quad}l}
\text{Input:}
&
\mathtt{\$}
\mathtt{X_1X_2\cdots X_t}\;
\underbrace{\sqcup\;\cdots\;\sqcup}_{g\text{ spaces}}\;
\mathtt{\$}
\\[3mm]
\text{Output:}
&
\mathtt{X_1X_2\cdots X_t}
\end{array}
\]
where $g$ is the gap between the source and target sequence.
Each $\mathtt{X_i}$ is a token sampled from $\{0, 1, \dots, 9\}$ uniformly at random. We can adjust $t, g$ to change the properties of this task. We use $t=5$ and $g\in\{64,8192\}$ for the training and evaluation in following experiments.

\subsection{Layer Retrieval Ability Score Estimation}
\label{sec:layer-contribution}

Here, we design a method for estimating a layer's contribution to the mode's retrieval abilities, which is important for us to monitor the learning of Spaced Retrieval of different layers in a hybrid model. Figure~\ref{fig:layer_contribution} provides an illustration of this method.

\paragraph{Model Notations}

Let $\mathbf{H}_{l-1}(x)\in\mathbb R^{N\times d}$ be the residual stream entering the $l$-th layer for input $x$. Each layer computes
\begin{align}
\mathbf{U}_l(x)
&= \operatorname{Mixer}_l
   \bigl(\mathbf{H}_{l-1}(x)\bigr),\\
\mathbf{R}_l(x)
&= \mathbf{H}_{l-1}(x)+\mathbf{U}_l(x),\\
\mathbf{H}_l(x)
&= \mathbf{R}_l(x)
 + \operatorname{MLP}_l
   \bigl(\mathbf{R}_l(x)\bigr),
\end{align}
where $\operatorname{Mixer}_l$ is a recurrence or softmax attention layer.
We write $\mathbf{h}_l(x,t)=[\mathbf{U}_l(x)]_t$ for its output
at token position $t$, before residual addition.
Let $\mathbf{z}(x,t)$ denote the final logits from the model's
native output head, and define the answer margin of two tokens $a, b$ as
\begin{equation}
m(\mathbf{z};a,b)=[\mathbf{z}]_a-[\mathbf{z}]_b.
\end{equation}

\paragraph{Retrieval Ability Estimation}
Consider two input prefixes $x^A$ and $x^B$ of equal
length $j$, differing only in the source digit to be retrieved: $a$ in $x^A$ and $b$ in $x^B$.
Their mixer outputs at layer $l$ are denoted by
$\mathbf{U}_l(x^A)$ and $\mathbf{U}_l(x^B)$.
We replace only the output at the last unmasked position. We also omit the layer index for simplicty, and we get:
\begin{equation}
\begin{aligned}
\mathbf{U}(x^A)
&=
\bigl[
\mathbf{U}_{1}(x^A),\ldots,
\mathbf{U}_{j-1}(x^A),\mathbf{U}_{j}(x^A)
\bigr],\\
\widetilde{\mathbf{U}}^{A\leftarrow B}
&=
\bigl[
\mathbf{U}_{1}(x^A),\ldots,
\mathbf{U}_{j-1}(x^A),\mathbf{U}_{j}(x^B)
\bigr],
\end{aligned}
\end{equation}
where $\mathbf{U}_{t}(x)$ denotes the mixer output at position $t$.

During the forward pass of $x^A$, we substitute
$\widetilde{\mathbf{U}}^{A\leftarrow B}$ for $\mathbf{U}(x^A)$
before residual addition, retaining the residual input from $x^A$.
The current layer's MLP and all subsequent computations proceed
normally with unchanged parameters.
We denote the resulting logits at position $j$ by
$\widetilde{\mathbf{z}}_{l,j}^{A\leftarrow B}$.



The \textbf{retrieval ability} score of layer $l$ for this input pair and answer position is
\begin{equation}
\boxed{
C_{l}^{A\leftarrow B}
=
m\bigl(\mathbf{z}(x^A, j);a,b\bigr)
-
m\bigl(\widetilde{\mathbf{z}}_{l,j}^{A\leftarrow B};a,b\bigr).
}
\end{equation}
This score indicates the reduction in correct-answer preference after replacing the layer's output with a misleading output with a different answer.
Therefore, the layers that contribute more to retrieval will have large scores, because the model's retrieval performance drops more severely when these layers' retrieval are interfered with.

\subsection{Analysis Experiments}
\label{sec:analysis-results}

With the above task and estimation method, we can now perform a series of experiments to analyze the phenomenon of \KEY{} for hybrid models, which is central to our paper.

\paragraph{Experimental Configuration}


We train several hybrid models on the Spaced Retrieval task. Each model has 3.6M parameters and consists of two KDA~(a widely-used linear recurrence architecture) layers and two attention layers. Unless specified otherwise, we use 1:1 mixtures of examples with $g=64$ and $g=8192$. During training, we only compute the loss on output tokens using loss masks. Here, we report the results for $g=8192$, but we emphasize that the results of $g=64$ are similar. The results for $g=64$ and more experimental details can be found in Appendix~\ref{sec:analysis-experimental-configuration}.

\subsubsection{Suppressing Learning in One Token Mixer Stimulates Learning in Another}
\label{sec:freeze_leading}

\begin{figure}[t]
    \centering

    \begin{minipage}[t]{0.48\linewidth}
        \centering
        \includegraphics[width=\linewidth]
        {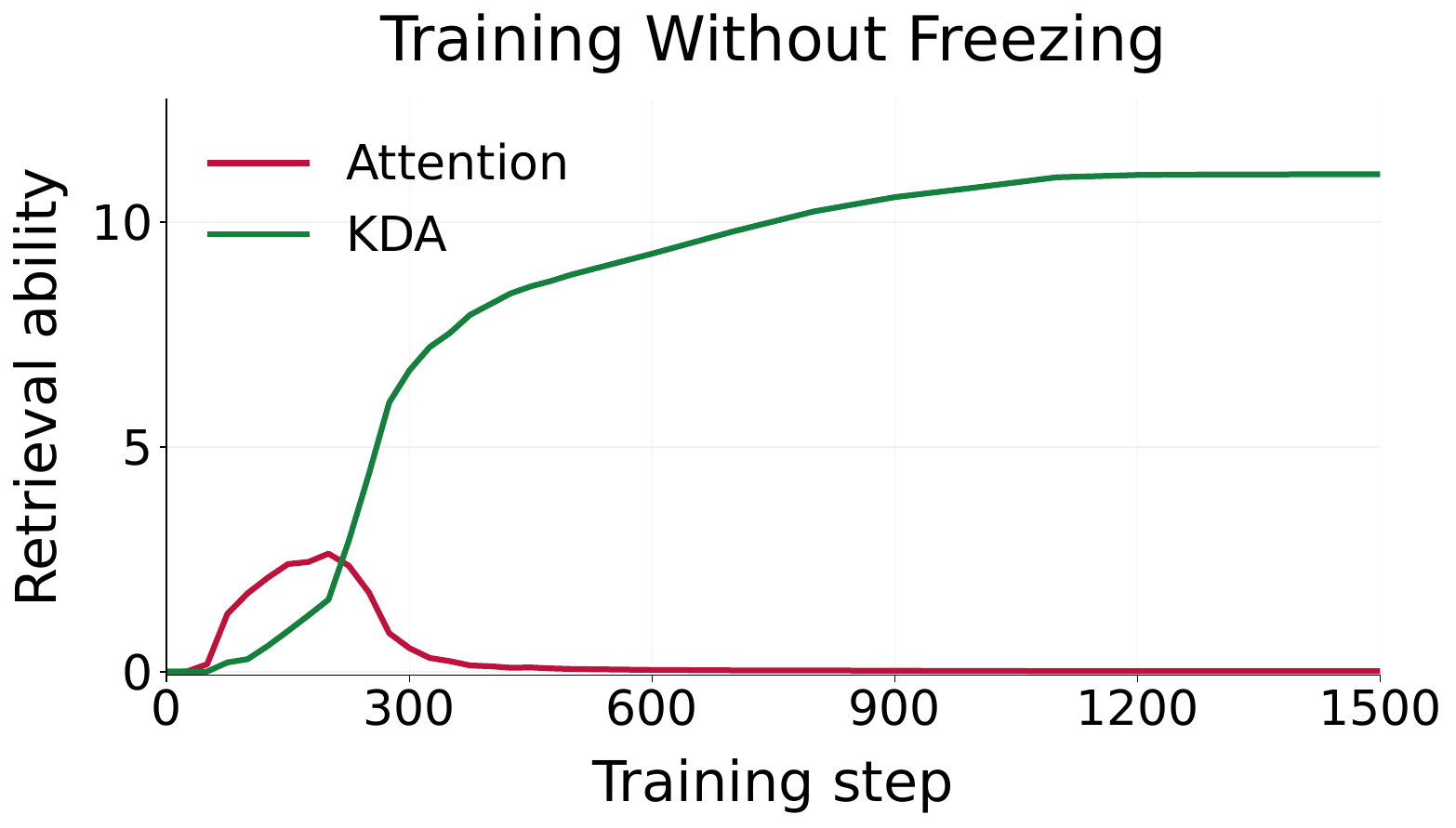}
        \caption{The average retrieval ability of the two types of layer
        in a hybrid model trained on Spaced Retrieval.}
        \label{fig:wild_compete}
    \end{minipage}
    \hfill
    \begin{minipage}[t]{0.48\linewidth}
        \centering
        \includegraphics[width=\linewidth]
        {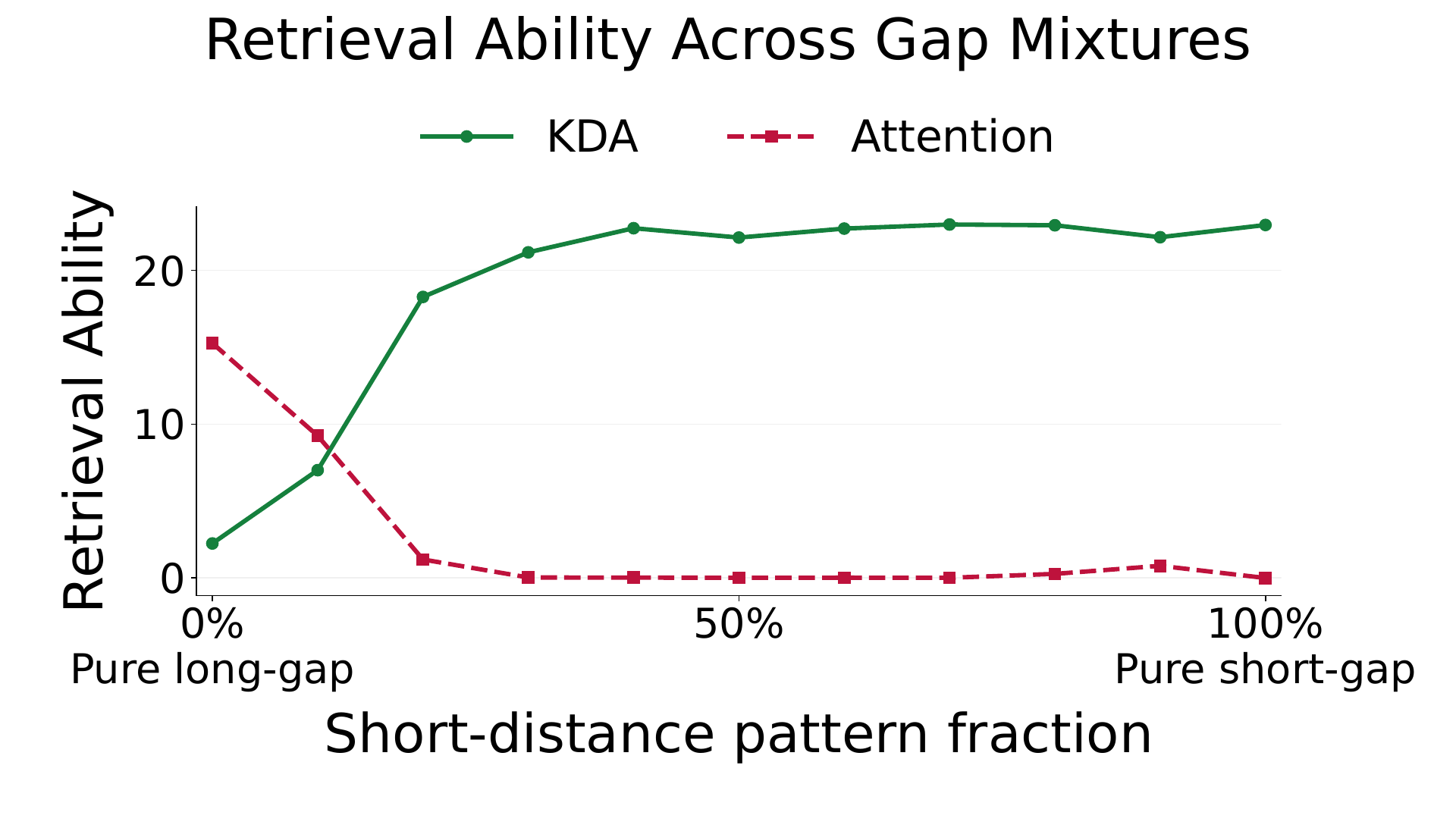}
        \caption{The retrieval ability of the two different kinds of layer
        in a hybrid model, trained on Spaced Retrieval with different data mixtures.}
        \label{fig:fraction_speed}
    \end{minipage}

\end{figure}

\begin{figure}[t]
    \centering
    \includegraphics[width=\linewidth]{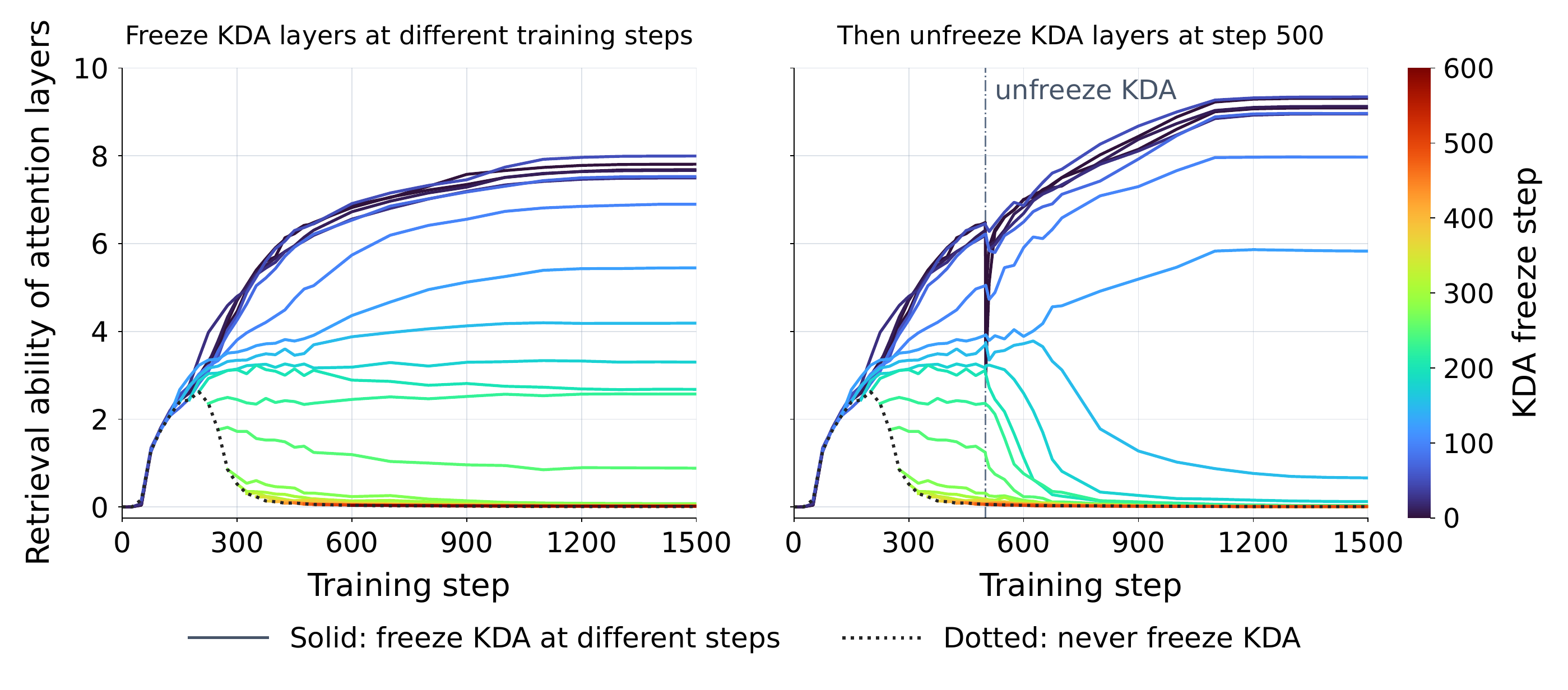}
    \caption{Demonstration of \KEY{}. The curves indicate the retrieval ability of attention layers in a hybrid model consisting of attention and recurrent layers~(KDA), trained on our retrieval task, Spaced Retrieval~(Section~\ref{sec:retrieval-task}).}
    \label{fig:freeze_leading}
\end{figure}


Without intervention, KDA layers become dominant in terms of \metric{} in this setting, as shown in Figure~\ref{fig:wild_compete}. To investigate how their dominance affects the attention layers, we freeze all KDA layers at different training steps and continue training the remaining model components. As shown in the left part of Figure~\ref{fig:freeze_leading}, the attention layers' \metric{} increases sharply immediately after each freezing intervention. This observation is consistent with the hypothesis that, once KDA layers acquire some retrieval capability, they suppress the acquisition of the same capability by attention layers, providing evidence for the \KEY{} phenomenon.

Furthermore, we explore how unfreezing KDA layers at a later training step influences learning competition. In the right plot of Figure~\ref{fig:freeze_leading}, runs in which KDA is frozen earlier than step 125 have higher retrieval ability than when the KDA remains frozen for the rest of the training, while runs in which KDA is frozen later than step 125 have lower retrieval ability than when the KDA remains frozen. Moreover, these changes appear almost immediately after KDA is unfrozen. This clear polarization may be attributed to the fact that attention layers in checkpoints where freezing began earlier than step 125 have enough retrieval advantages by freezing KDA layers long enough.
At these early freezing checkpoints, attention layers have already accumulated a sufficient advantage in retrieval performance, substantially outperforming KDA layers. Consequently, once freezing is lifted, attention layers tend to further reinforce their retrieval capabilities, while KDA layers are driven to specialize in other functions. This is reflected in Figure~\ref{fig:freeze_leading}: at checkpoints where attention layers have established a sufficiently strong advantage, their retrieval ability further improves after unfreezing. In contrast, at checkpoints where this advantage is insufficient, attention layers are once again outcompeted by KDA layers, leading to a decline in their retrieval ability.

\paragraph{Takeaways} In hybrid models, KDA and attention layers compete to learn retrieval, with KDA initially gaining an advantage due to its stronger inductive bias toward short-range patterns. Temporarily freezing the dominant KDA layers can shift---and, with sufficient duration, even reverse---the winner of this competition.

\subsubsection{Gradient Evidence of \KEY}
\label{sec:grad_change}


\begin{figure}[t]
    \centering
    \includegraphics[width=1\linewidth]{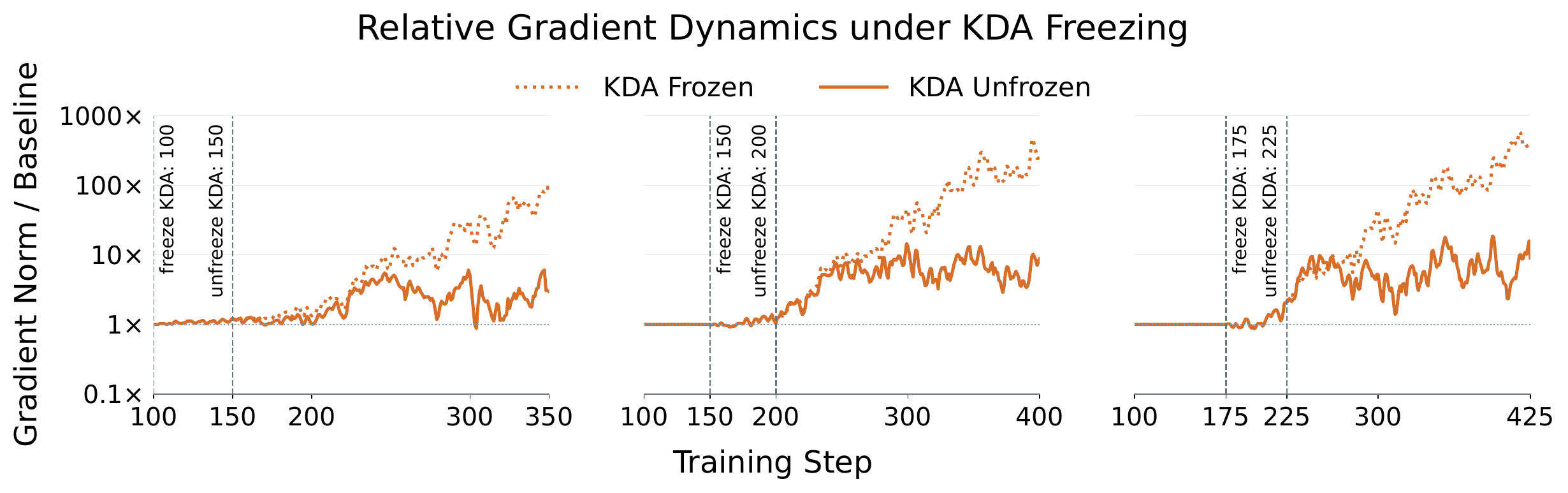}
    \caption{Relative gradient magnitude changes in the Attention layers compared to the baseline without KDA freezing. KDA is frozen at steps 100, 150, or 175, with an unfreezing branch introduced 50 steps later. Across these timings, keeping KDA frozen progressively amplifies attention gradient norms relative to the baseline, whereas unfreezing KDA substantially attenuates this amplification.}
    \label{fig:grad_change}
\end{figure}

We further investigate how the freezing intervention affects gradient dynamics. As shown in Figure~\ref{fig:grad_change}, the gradient norm of the attention layers in the frozen checkpoint reaches approximately $10\times$ that of the baseline checkpoint, which has never undergone freezing. Moreover, the gap between the dotted and solid curves indicates that unfreezing the KDA layers mitigates this gradient amplification. Notably, the gradient norm increases progressively rather than abruptly, suggesting that this amplification emerges from the evolving optimization dynamics rather than as an immediate consequence of freezing. This observation further supports \KEY, suggesting a self-reinforcing optimization mechanism in which the current parameter configuration determines the allocation of learning signals, progressively amplifying the advantages of already dominant components. See Appendix~\ref{app:grad-change} for more details.

\paragraph{Takeaways} Freezing KDA progressively amplifies attention layers' gradients, while unfreezing mitigates this effect. This supports \KEY: learning signals depend on the current ability gap of components, reinforcing the advantages of already dominant components.

\subsubsection{Different Training Data Prefer Different Architectures}
\label{sec:fraction_speed}


\begin{figure}[t]
    \centering

    \begin{minipage}[t]{0.48\linewidth}
        \centering
        \includegraphics[width=\linewidth]{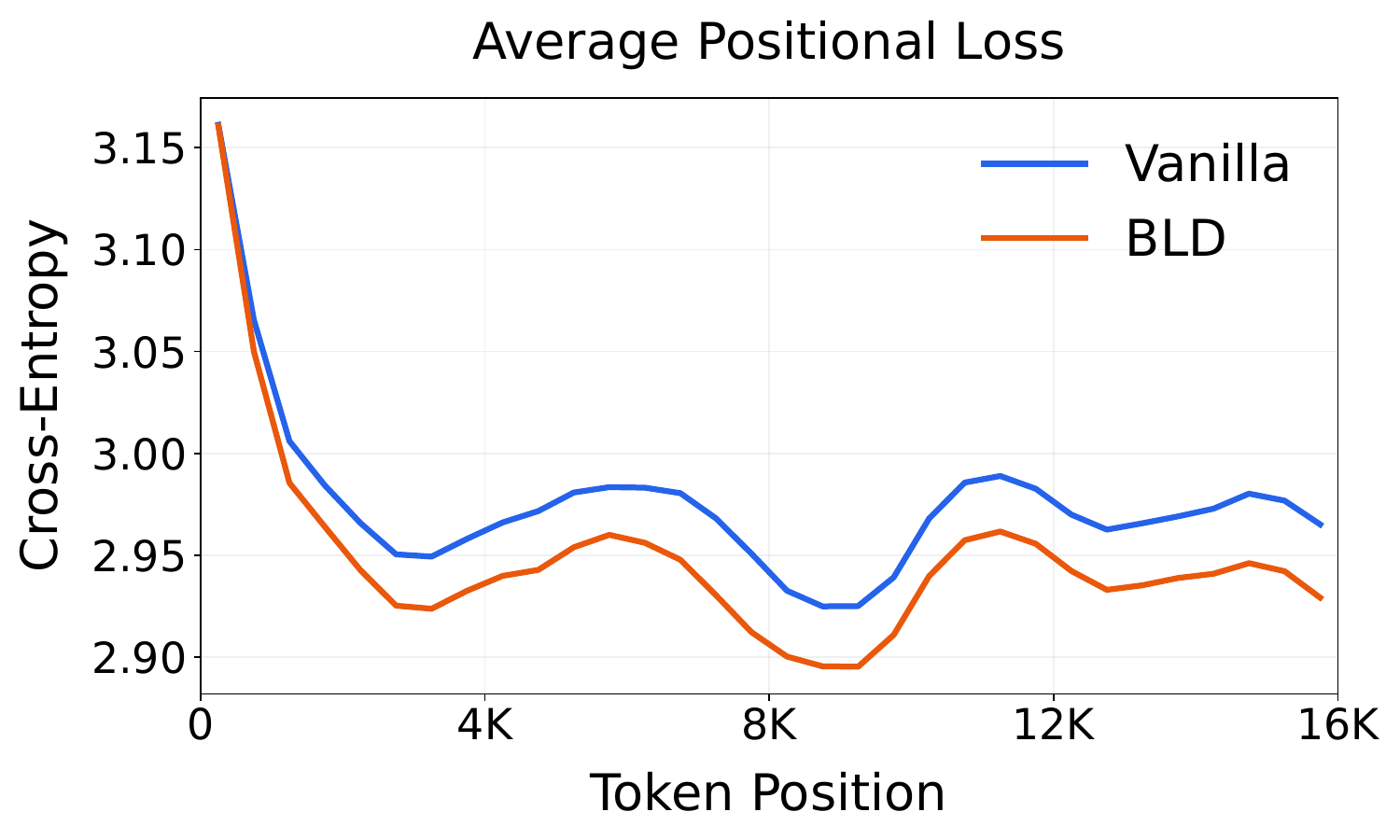}
        \caption{Average position-wise loss of 510M \OURS/vanilla model across all examples.}
        \label{fig:long_loss}
    \end{minipage}
    \hfill
    \begin{minipage}[t]{0.48\linewidth}
        \centering
        \includegraphics[width=\linewidth]{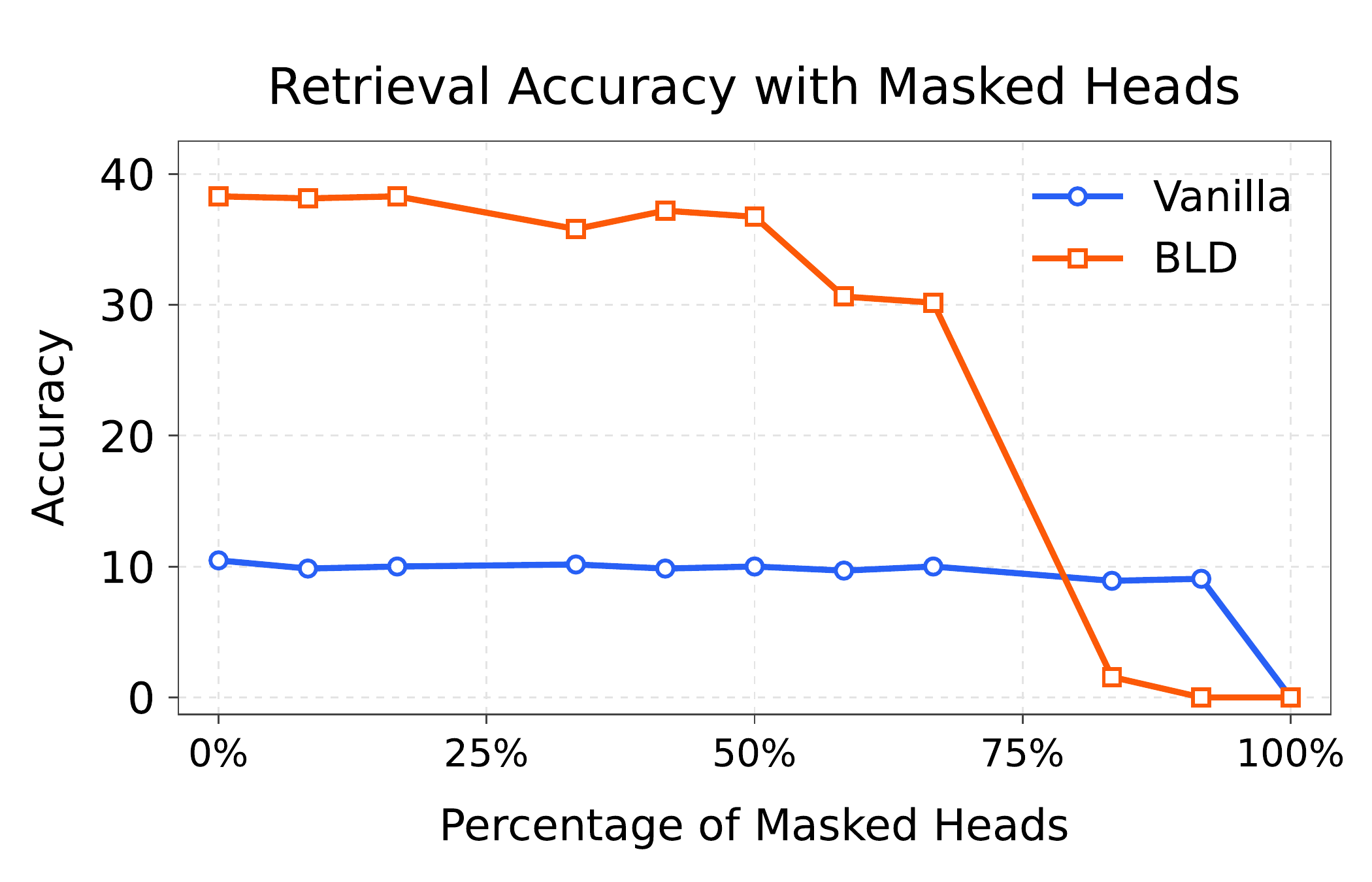}
        \caption{The retrieval accuracy when masking a fraction of the heads.}
        \label{fig:mask_acc}
    \end{minipage}

\end{figure}


We further investigate how the distribution of gap $(g)$ in the training data influences the learnability of retrieval in different mixers.
To this end, we train the hybrid model using different mixture proportions of short-distance ($g=64$) and long-distance ($g=8192$) examples. As shown in Figure~\ref{fig:fraction_speed}, attention layers become more important as the proportion of long-distance examples increases, whereas KDA layers are more likely to dominate when the long-distance patterns do not appear overwhelmingly more than short-distance patterns. This is because KDA initializes forget gates with low values, which gives this architecture a short-distance inductive bias (See Appendix~\ref{app:forget-gate} for a more specific explanation).

\paragraph{Takeaways}

Different mixers learn retrieval at different speeds, depending on the data distribution. While both mixers can acquire the same retrieval abilities, the mixer that learns faster will become dominant, limiting the learning of the other mixer.


\paragraph{Implications}

Applying the traditional training recipe to a hybrid model can lead to a suboptimal division of labor. Because short-distance patterns occur more frequently than long-distance patterns in natural language~\citep{doi:10.1073/pnas.1502134112}, the training objective may encourage KDA to dominate retrieval, suppressing the training of attention layers. This issue is less problematic in homogeneous architectures such as Transformers, where all layers share the same architecture and, consequently, have the same potential for retrieval.


\section{LLM Pretraining Experiments}

In this section, we show that we can leverage insights about \KEY{} to achieve improved LLM performance in natural language pre-training.

\subsection{Method: Better Layer Division}


Based on the analysis results in Section \ref{sec:freeze_leading}, we propose Better Layer Division~(BLD), which freezes RNN layers' weights during the early stage of pretraining, thereby giving the attention layers a first-mover advantage in \KEY. Importantly, this intervention establishes a corrected division of labor between the two components, which is subsequently preserved and further reinforced throughout the remainder of pretraining. This better labor division translates to improved model performance.

\subsection{Experimental Details}
\label{sec:experimental-details}





\paragraph{Model Configuration}
We train a family of KDA--attention hybrid models ranging from
180M to 1B total parameters, including embeddings.
All models share a repeating four-layer pattern consisting of
one attention layer followed by three KDA layers, maintaining
a fixed 3:1 KDA-to-attention ratio across scales.
The layers are stacked in the form of ``Attention-KDA-KDA-KDA'', and all layers are using NoPE (not adopting RoPE, which is reported to be better than using RoPE in KDA-based hybrid architecture~\citep{team2025kimi}).
We scale the models by varying their depth and hidden width.
At each scale, \OURS and vanilla pre-training use identical
architectures.
Detailed model configurations are provided in
Appendix~\ref{app:scaling_model_details}.

\paragraph{Training Configuration}
We compare \OURS against vanilla pre-training by training
models from scratch with token budgets scaled proportionally
to their non-embedding parameter counts, following the
Chinchilla scaling principle (training tokens are roughly 20 $\times$ of parameters) ~\citep{hoffmann2022training}.
All models are trained on DCLM~\citep{li2024datacomplm}, a high-quality open-source pre-training corpus,
using the Qwen3 tokenizer~\citep{yang2025qwen3} and a context length of 16,384 tokens.
For \OURS, the KDA blocks remain frozen for approximately $1/10$  of the subsequently unfrozen pretraining part.
The vanilla baseline updates all parameters,
using the same training data, token budgets, and optimization
schedule.


We use AdamW during the initial portion of pre-training,
followed by Muon with auxiliary AdamW after unfreezing,
with identical optimization schedules for \OURS and the
vanilla baseline.
Detailed hyperparameters and token budgets are provided
in Appendix~\ref{app:scaling_training_details}.

\paragraph{Evaluation Tasks}

We evaluate models across common-sense reasoning~(CSR), recall-intensive tasks, RULER, and language modeling, without fine-tuning. See Appendix~\ref{app:scaling_evaluation_details} for more details.

\begin{itemize}
    \item CSR is measured by the average score on seven benchmarks: ARC-Easy, ARC-Challenge~\citep{allenai:arc}, HellaSwag~\citep{zellers2019hellaswag}, PIQA~\citep{Bisk2020}, BoolQ~\citep{clark2019boolq}, WinoGrande~\citep{ai2:winograde}, LAMBADA~\citep{paperno2016lambada}. 
    \item Recall is evaluated on six tasks from the Based~\citep{arora2024simple} evaluation suite: SWDE~\citep{lockard2019openceres}, FDA~\citep{arora2023language}, SQuAD~\citep{rajpurkar2018know}, TriviaQA~\citep{joshi2017triviaqa}, DROP~\citep{dua2019drop}, and Natural Questions~\citep{kwiatkowski2019natural}. 
    \item RULER~\citep{hsieh2024ruler} is evaluated with average scores across the 57 evaluated RULER task–length combinations spanning 1K to 16K tokens. 
    \item Language modeling is evaluated using word-level perplexity on WikiText-2~\citep{merity2016pointer} and selected-token perplexity on LongPPL~\citep{fang2025wrong}. 
\end{itemize}

\begin{table*}[!t]
    \centering
    \scriptsize
    \begin{tabular}{cl|c|ccc|cc}
    \toprule
    Size & Model & LM Loss $\downarrow$
    & CSR $\uparrow$ & Recall $\uparrow$ & RULER $\uparrow$
    & WikiText $\downarrow$ & LongPPL $\downarrow$ \\
    \midrule
    180M & Vanilla & 3.5903 & \textbf{37.76} & 10.39 & 2.84 & 54.00 & 258.47 \\
    \rowcolor{orange!10}
    180M & \OURS & \textbf{3.5702} & 37.49 & \textbf{14.12} & \textbf{8.82} & \textbf{50.38} & \textbf{108.64} \\
    \midrule
    250M & Vanilla & 3.2844 & 38.31 & 21.54 & 17.32 & 36.52 & 57.22 \\
    \rowcolor{orange!10}
    250M & \OURS & \textbf{3.2642} & \textbf{41.49} & \textbf{25.33} & \textbf{20.98} & \textbf{34.27} & \textbf{17.96} \\
    \midrule
    390M & Vanilla & 3.1611 & 41.91 & 26.73 & 22.32 & 29.10 & 17.09 \\
    \rowcolor{orange!10}
    390M & \OURS & \textbf{3.1441} & \textbf{44.25} & \textbf{31.27} & \textbf{26.89} & \textbf{28.11} & \textbf{11.16} \\
    \midrule
    510M & Vanilla & 3.0553 & \textbf{44.80} & 30.50 & 25.25 & 26.02 & 14.56 \\
    \rowcolor{orange!10}
    510M & \OURS & \textbf{3.0461} & 44.49 & \textbf{34.70} & \textbf{29.42} & \textbf{25.57} & \textbf{9.21} \\
    \midrule
    750M & Vanilla & 2.9773 & 45.94 & 33.52 & 26.53 & 23.28 & 12.68 \\
    \rowcolor{orange!10}
    750M & \OURS & \textbf{2.9705} & \textbf{46.61} & \textbf{37.52} & \textbf{32.82} & \textbf{22.75} & \textbf{8.08} \\
    \midrule
    1B & Vanilla & 2.8744 & 48.69 & 36.50 & 32.75 & 20.52 & 7.69 \\
    \rowcolor{orange!10}
    1B & \OURS & \textbf{2.8680} & \textbf{49.45} & \textbf{38.14} & \textbf{36.04} & \textbf{20.15} & \textbf{6.50} \\
    \bottomrule
    \end{tabular}
    \caption{Zero-shot evaluation results of BLD~(ours) versus vanilla
    pre-training. Size denotes total parameters, including embeddings,
    rounded to the nearest 10M.
    CSR and Recall are macro averages over seven and six tasks, respectively.
    RULER averages 57 task--length entries at 1K--16K.
    All three scores are percentages.
    Bold indicates the better result at each model size. }
    \label{tab:downstream_result}
\end{table*}

\subsection{Main Results}

\paragraph{Downstream Tasks}
As shown in Table~\ref{tab:downstream_result}, \OURS models achieve notably lower LM losses. They also show a modest improvement in CSR performance, while exhibiting substantial gains on retrieval-oriented tasks, including Recall and RULER. Moreover, the PPL metrics, which reflect overall long-context modeling capability, consistently favor \OURS by a large margin. The specific subtask scores are included in Appendix~\ref{app:scaling_results}.

\paragraph{Context Utilization}
We evaluate the long-context utilization of our models as \citet{sun2024learning} by computing LM loss as a function of context length. Specifically, we calculated next-token prediction loss on \texttt{Books3,BookCorpus2}~\citep{gao2020pile} as a function of token positions, up to 16K. Figure~\ref{fig:long_loss} shows that \OURS model has a lower loss especially in the later position, highlighting better context utilization abilities by \OURS models. The results of all model sizes and other details are included in Appendix~\ref{app:long_loss}.

\begin{figure}[t]
    \centering
    \includegraphics[width=\linewidth]{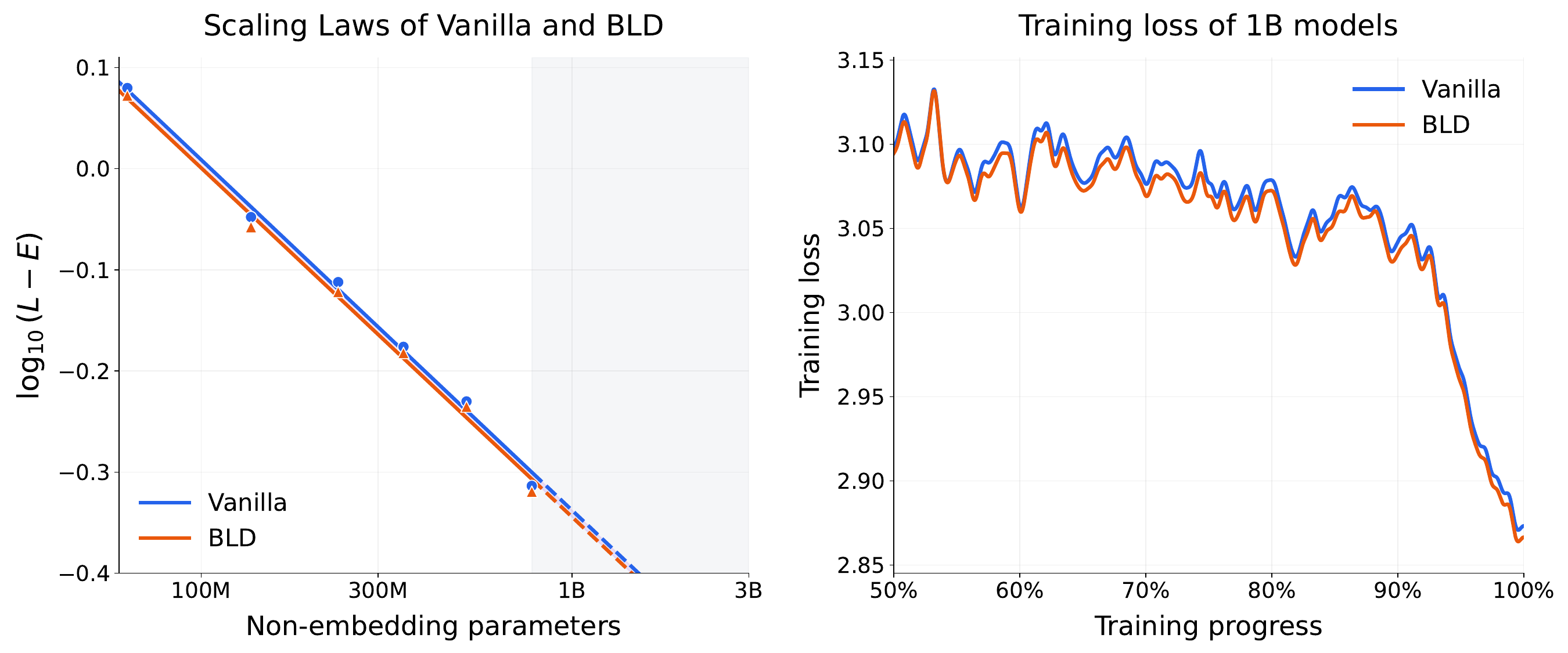}
    \caption{Left: Loss as a function of non-embedding model size, fitted with power-plus-constant scaling laws. Right: The training-loss curve of the 1B vanilla and BLD models.}
    \label{fig:scaling_laws}
\end{figure}

\paragraph{Scaling Laws}
We also fit the loss as a power-law function of model size to verify that \OURS is scalable to larger model scales~(shown in Figure~\ref{fig:scaling_laws}), and the fitted curves suggest that \OURS retains its loss advantage well beyond the evaluated model scales.

\subsection{Transferability to Other RNN Architectures}
\label{sec:transfer-to-other-architectures}

\begin{table*}[t]
    \centering
    \scriptsize
    \setlength{\tabcolsep}{4pt}
    \begin{tabular}{ccl|c|ccc|cc}
    \toprule
    Architecture & Size & Model & LM Loss $\downarrow$
    & CSR $\uparrow$ & Recall $\uparrow$ & RULER $\uparrow$
    & WikiText $\downarrow$ & LongPPL $\downarrow$ \\
    \midrule
    GDN & 400M & Vanilla & 3.1947 & 43.23 & 17.33 & 12.33 & 31.25 & 152.84 \\
    \rowcolor{orange!10}
    GDN & 400M & \OURS & \textbf{3.1624} & \textbf{43.90} &\textbf{ 32.05} & \textbf{25.42} & \textbf{28.69} & \textbf{11.77} \\
    \midrule
    GLA & 400M & Vanilla & 3.2911 & 41.19 & 24.58 & 23.98 & 34.57 & 32.69 \\
    \rowcolor{orange!10}
    GLA & 400M & \OURS & \textbf{3.2113} & \textbf{43.20} & \textbf{30.12} & \textbf{28.54} & \textbf{30.65} & \textbf{13.26} \\
    \bottomrule
    \end{tabular}
    \caption{Zero-shot evaluation results of BLD~(ours) versus vanilla
    pre-training with GDN and GLA hybrid models with 3:1 recurrence-to-attention ratio.}
    \label{tab:gdn_gla_downstream_result}
\end{table*}

We also applied \OURS to two other popular recurrent mixers, Gated DeltaNet~(GDN)~\citep{yang2025gated} and Gated Linear Attention~(GLA)~\citep{yang2023gated}. 
These models have the same configuration as the model in Section~\ref{sec:experimental-details}, but with KDA layers replaced with either of these layers.
Training configuration is also the same as in Section~\ref{sec:experimental-details}. Further details are reported in Appendix~\ref{app:gdn_and_gla}.

\paragraph{Takeaways}

The advantages of \OURS, lower training loss and better downstream task performance, are transferable to other linear recurrence architectures.

\subsection{Evidence for Labor Division Change in Attention Scores}
\label{sec:retrieval-head}

\begin{figure}[t]
    \centering
    \includegraphics[width=\linewidth]{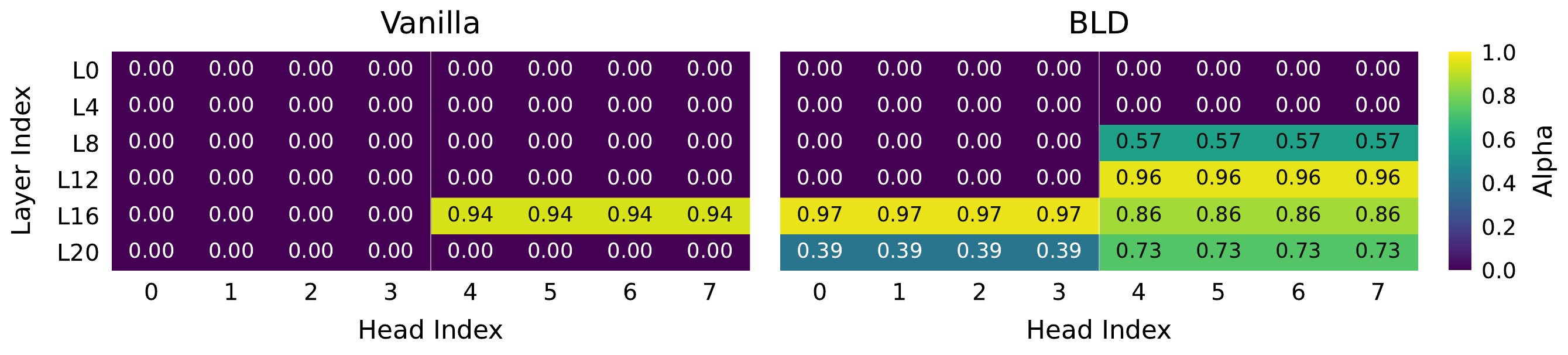}
    \caption{The $\alpha$ values of attention heads. With GQA, query heads 0–3 and 4–7 share a KV head within each group and therefore share the same KV-head-level $\alpha$ value. Greater $\alpha$ values indicate heads that are important for long-distance retrieval.}
    \label{fig:alpha_values}
\end{figure}


We further provide direct evidence that \OURS{} induces stronger functional differentiation. Specifically, following \citet{xiao2025duoattention}, we analyze 510M KDA-attention hybrid models by training \textit{$\alpha$ values} to distinguish \textit{retrieval heads}, which are responsible for long-distance information retrieval, from \textit{streaming heads}, which primarily perform short-distance token mixing. 
$\alpha$ measures a head’s reliance on full-context attention, with larger values indicating greater reliance. In other words, heads with greater $\alpha$ values are more important for long-distance retrieval. The $\alpha$ values are trained for 1,000 steps. More experimental details and results are reported in Appendix~\ref{app:alpha-value}.



As shown in Figure~\ref{fig:alpha_values}, the attention heads in the \OURS{} model exhibit higher $\alpha$ values than those in the vanilla model, indicating that \OURS{} contains more retrieval heads that attend to long-range tokens. Figure~\ref{fig:mask_acc} reports model's retrieval accuracy as attention heads are progressively masked in ascending order of their $\alpha$ values. For \OURS{}, the accuracy remains largely unchanged as low-$\alpha$ heads are masked and drops sharply only when the final three heads, which have the highest $\alpha$ values, are removed. \textbf{This suggests that these three heads play a critical role in retrieval, and we therefore identify them as retrieval heads. In contrast, the vanilla model exhibits a substantial accuracy drop only when its single highest-$\alpha$ head is masked, indicating that it contains only one retrieval head.} 

\paragraph{Takeaways}

\OURS{} promotes a greater degree of functional specialization, with more attention heads specializing in long-range retrieval. This behavior aligns with the motivation underlying our method.

\section{Conclusion}


In this work, we first identify the \KEY phenomenon and provide supporting experimental evidence. We then show that directly pretraining hybrid models can lead to a suboptimal division of labor among different architectural components. To address this issue, we propose \OURS and validate its effectiveness through scalable experiments. By promoting a more effective division of labor, \OURS consistently improves model capabilities across a wide range of dimensions.

Nevertheless, \OURS represents only a preliminary approach to addressing the \KEY phenomenon in the training of hybrid models. We believe that a deeper understanding of the \KEY phenomenon will provide further insights into the division of labor within hybrid architectures and inspire the development of more advanced neural architectures and training strategies that better account for this phenomenon and achieve stronger performance.

\subsection*{AI use statement}

In this work, we used generative AI tools for research execution.
We have not used generative AI tools for research ideation,
and LLM data-generation is not applicable to this work.
Additionally, we used generative AI tools for polishing the manuscript. We have reviewed all AI-assisted work. LLM-generated code was verified and tested for correctness by 2 authors. We take responsibility for the final content of this work,
including text, claims or artifacts produced with the aid of generative AI.




\subsection*{Reproducibility statement}

We provide the experimental settings needed to reproduce our results
in the appendix. Appendix~\ref{sec:analysis-experimental-configuration}
details the synthetic retrieval task, model configuration, training
procedure, and freezing interventions.
Appendices~\ref{app:scaling_model_details}
and~\ref{app:scaling_training_details} describe the language-model
architectures, training data, token budgets, optimization settings,
and parameter-freezing scope.
Evaluation protocols and detailed results are provided in
Appendices~\ref{app:scaling_evaluation_details}
and~\ref{app:scaling_results}.
Appendices~\ref{app:long_loss} and~\ref{app:gdn_and_gla}
describe the position-wise loss analysis and the GDN/GLA experiments,
respectively.

We will publicly release the training and evaluation code, relevant
experimental data, and model checkpoints to facilitate reproduction.



\bibliography{iclr2027_conference}
\bibliographystyle{iclr2027_conference}

\clearpage
\appendix
\clearpage
\section{Spaced Retrieval Experiments}
\label{sec:analysis-experimental-configuration}

\subsection{Data and architecture}
We use the Spaced Retrieval task from Section~\ref{sec:retrieval-task}:
\[
\begin{array}{r@{\quad}l}
\text{Input:} &
\mathtt{\$}\mathtt{X_1X_2\cdots X_t}\;
\underbrace{\sqcup\;\cdots\;\sqcup}_{g\text{ spaces}}\;\mathtt{\$}
\\[3mm]
\text{Output:} & \mathtt{X_1X_2\cdots X_t}
\end{array}
\]
Here $t=5$, each $X_i$ is a digit, and $g\in\{64,8192\}$ is the number of
spaces separating the key from the retrieval query.
We add 0–20 spaces before the first delimiter to vary the key’s position. Training and evaluation, including the modified keys used for attribution,  use disjoint keys. To compare \(g=64\) and \(g=8192\), we keep each example identical except for \(g\).
Tables~\ref{tab:app-toy-data} and~\ref{tab:app-toy-training} summarize the data, model, and optimization settings.

\begin{table}[!h]
    \centering\small
    \renewcommand{\arraystretch}{1.12}
    \caption{The configuration of the dataset~(Spaced Retrieval) and the models in Section~\ref{sec:analysis-results}.}
    \label{tab:app-toy-data}
    \begin{tabular}{@{}p{0.29\linewidth}p{0.65\linewidth}@{}}
    \toprule
    Setting & Value \\
    \midrule
    Key length / $g$ & Five digits; $g\in\{64,8192\}$ \\
    Leading spaces & 0--20 \\
    Vocabulary & 12 symbols: ten digits, delimiter, and space/padding \\
    Training / evaluation & 50,000 distinct keys / 2,000 held-out pairs per $g$ \\
    Decoder blocks & KDA--attention--KDA--attention \\
    Hidden / FFN width & 256 / 512 \\
    Mixer heads / attention KV heads & 2 / 2 \\
    KDA head dimension / value expansion & 128 / 2 \\
    Short convolution width & 4 \\
    KDA output gate & Enabled (sigmoid)~\citep{team2025kimi} \\
    position encoding & all layers have no rotary embeddings \\
    Embeddings & Tied input/output embeddings \\
    \bottomrule
    \end{tabular}
\end{table}

\subsection{The training configuration of the experiments in Section~\ref{sec:analysis-results}.}

\begin{table}[!h]
    \centering\small
    \renewcommand{\arraystretch}{1.12}
    \caption{Synthetic training and attribution evaluation settings.}
    \label{tab:app-toy-training}
    \begin{tabular}{@{}p{0.29\linewidth}p{0.65\linewidth}@{}}
    \toprule
    Setting & Value \\
    \midrule
    Optimizer / updates & AdamW~\citep{loshchilov2019decoupled} / 1,500 \\
    Batch & 32 examples per update \\
    Learning rate & Peak $3\times10^{-4}$; final $10^{-5}$ \\
    Schedule & 100 warmup updates, followed by cosine decay \\
    Adam coefficients & $(\beta_1,\beta_2)=(0.9,0.95)$ \\
    Weight decay / gradient clip & 0.1 / global norm 1.0 \\
    Precision & BF16 \\
    Loss mask & Cross-entropy on the five answer tokens only \\
    \bottomrule
    \end{tabular}
\end{table}

Source digits, delimiters, leading spaces, and gap tokens do not contribute
to the training loss.

For attribution, we select one answer digit per held-out pair, with 400 pairs
assigned to each of the five answer positions. The counterfactual input changes
only the corresponding digit in the source key. Each input prefix ends just
before the selected answer digit, and any earlier answer digits are identical
in the two inputs. We average attribution scores across these pairs, rather
than performing five interventions on every pair.

\subsection{Bidirectional mixer-output intervention}
For each input pair, we compute the retrieval ability score defined in
Section~\ref{sec:layer-contribution} twice, once in each direction
($A\leftarrow B$ and $B\leftarrow A$), and take their average.
We then average over all evaluation pairs:
\begin{equation}
 C_l=\frac{1}{2}\mathbb{E}_{(A,B)}\!\left[
 m(\mathbf z^A;a,b)-m(\widetilde{\mathbf z}^{A\leftarrow B}_l;a,b)
 +m(\mathbf z^B;b,a)-m(\widetilde{\mathbf z}^{B\leftarrow A}_l;b,a)
 \right].
\end{equation}
In each direction, we replace the selected layer's mixer output at the
prediction position with that from the paired input, then recompute the
prediction without changing the model weights.

The score is a change in a \emph{logit margin}.
Signed values are retained. Independently measured layer effects need not add
to a total model effect: they are neither disjoint contributions nor normalized
shares. In the baseline and freeze/unfreeze plots, a mixer-family curve averages
its two layer scores. The mixture endpoint plot instead sums the two scores
within each family before averaging across seeds.

\subsection{Freezing and unfreezing interventions}
\begin{table}[htbp]
\centering\small
\renewcommand{\arraystretch}{1.12}
\caption{Synthetic freezing and unfreezing protocol for the experiment in Section~\ref{sec:freeze_leading}.}
\label{tab:app-toy-freezing}
\begin{tabular}{@{}p{0.29\linewidth}p{0.65\linewidth}@{}}
\toprule
Setting & Value \\
\midrule
Frozen parameters & Entire KDA blocks 0 and 2, including mixer, MLP, and normalization \\
Trainable components & Attention blocks and shared parameters \\
Updates completed before freezing ($s$) & 0, 1, 10, 25, 50, 75, 100, 125, 150, 175, 200, 225, 250, 275, 300, 325, 350, 375, 400, 425, 450, 475, 500, 600 \\
Permanent freeze & Keep KDA blocks frozen through update 1,500 \\
\bottomrule
\end{tabular}
\end{table}

For each $s$ in Table~\ref{tab:app-toy-freezing}, we train normally for $s$
updates, then freeze the KDA blocks starting with update $s+1$. Thus, $s=0$
means freezing from the start, and the score recorded at update $s$ is measured
before freezing has any effect. Frozen blocks still participate in the forward
pass and allow gradients to reach earlier trainable components.

In the unfreezing experiment, we resume KDA updates at update 501 and continue
training to update 1,500. We use only $s\leq500$, excluding $s=600$.
The $s=500$ run has no frozen updates and serves as a continuation control;
we also compare with the baseline that never freezes KDA.

Each run continues from the saved training state, preserving the optimizer,
learning-rate schedule, random state, and data order. Unfreezing does not
reset these states or restart warmup. 

\begin{figure}[htbp]
\centering
\includegraphics[width=\linewidth]{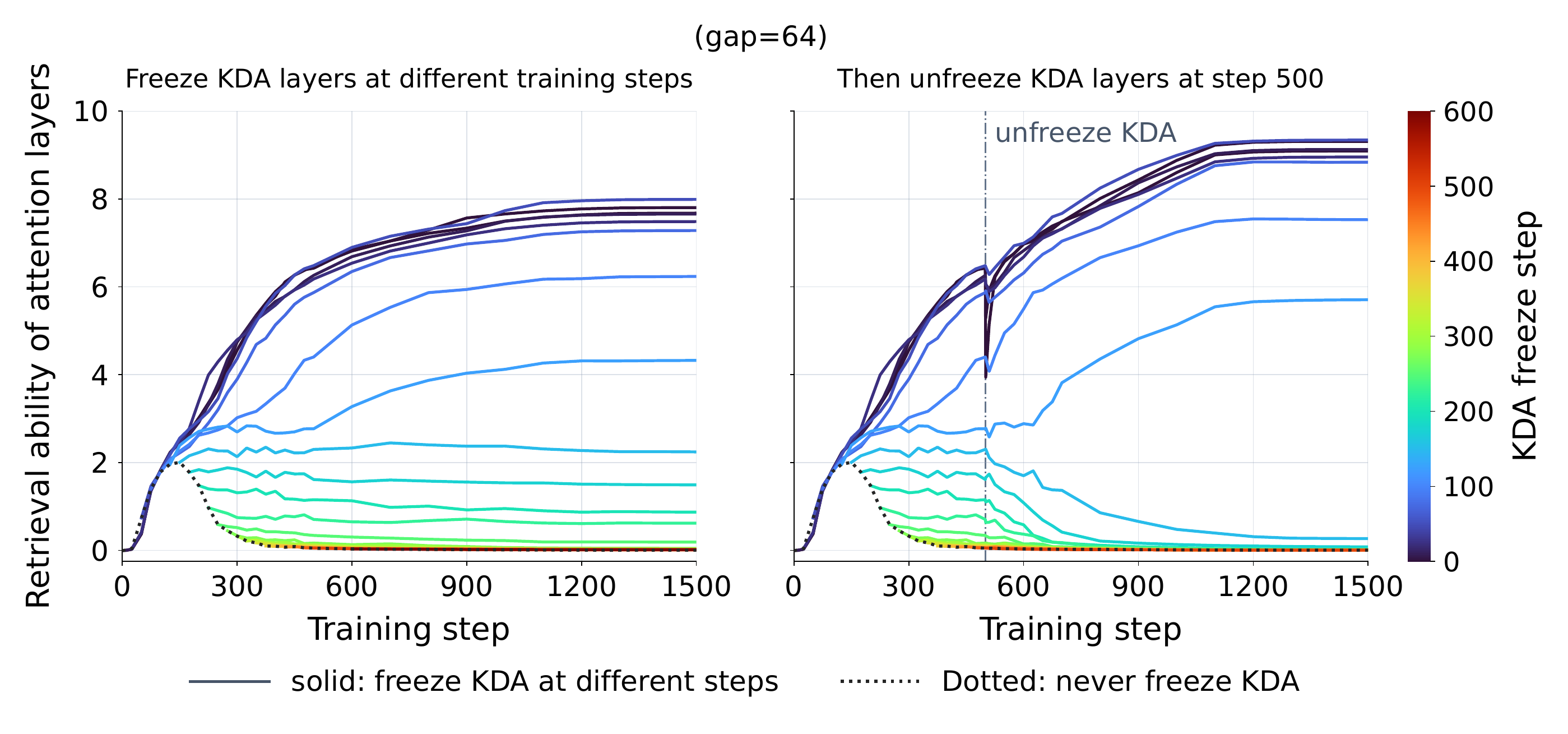}
\caption{Attention retrieval ability under KDA freezing and unfreezing at
$g=64$, using the same setup as in Figure~\ref{fig:freeze_leading}.}
\label{fig:app-short-freeze}
\end{figure}

\subsection{Different Training Data Prefer Different Architectures}
\label{appendix:detective_ablation}
We vary the proportion of short-distance examples to study how the training
distribution affects retrieval learning, as in Section~\ref{sec:fraction_speed}.
Let $p$ denote the fraction of examples with $g=64$; the remaining fraction
$1-p$ uses $g=8192$. Thus, $p=0$ uses only long-distance examples, whereas
$p=1$ uses only short-distance examples.

\begin{table}[htbp]
\centering\small
\renewcommand{\arraystretch}{1.12}
\caption{Training-data proportions and settings for the data-distribution experiment in Section~\ref{sec:fraction_speed}.}
\label{tab:app-mixture-setup}
\begin{tabular}{@{}p{0.29\linewidth}p{0.65\linewidth}@{}}
\toprule
Setting & Value \\
\midrule
Short-distance fraction $p$ & 0, 0.1, 0.2, 0.3, 0.4, 0.5, 0.6, 0.7, 0.8, 0.9, 1 \\
Training distances & $g=64$ and $g=8192$ \\
Optimizer updates & 1,500 \\
Examples per update & 32 \\
Evaluation distances & $g=64$ and $g=8192$ \\
\bottomrule
\end{tabular}
\end{table}

Within each optimizer update, short- and long-distance examples are processed
separately. Their gradients are weighted by the number of examples and
accumulated before a single parameter update. We use at least three random
seeds for each setting and report the average final-checkpoint retrieval
scores in Figure~\ref{fig:fraction_speed}.

\section{Short-distance Inductive Bias from KDA Forget Gates}
\label{app:forget-gate}

The KDA mechanism can be formulated as follows:
\begin{equation}
\begin{aligned}
    S_t &= (I - \beta_t k_t k_t^T) \operatorname{Diag}(\alpha_t) S_{t-1} + \beta_t k_t v_t^T \in \mathbb{R}^{d_k \times d_v}, \
    o_t &= S_t^T q_t \in \mathbb{R}^{d_v}.
\end{aligned}
\end{equation}
Here, $\alpha_t$ denotes the forget gate, which is initialized with relatively small values. Consequently, the KDA state $S_t$ initially tends to retain information primarily from recent tokens. This is because historical information is progressively attenuated by the forget gates, and the cumulative product of small gate values rapidly approaches zero over successive time steps. As a result, contributions from distant tokens diminish quickly.

\section{Model Configurations}
    \label{app:scaling_model_details}
    \subsection{Parameter accounting and model identity}


\begin{table}[!h]
    \centering
    \footnotesize
    \setlength{\tabcolsep}{5pt}
    \renewcommand{\arraystretch}{1.1}
    \caption{KDA hybrid model configurations. Parameter counts
    are rounded to the nearest million; tied embeddings are
    counted once.}
    \label{tab:app-architecture}
    \begin{tabular}{l|rr|rrrrr}
        \toprule
        \shortstack{Model\\size} &
        \shortstack{Total\\params (M)} &
        \shortstack{Non-emb.\\params (M)} &
        \shortstack{Decoder\\blocks} &
        \shortstack{Hidden\\dim.} &
        \shortstack{FFN\\dim.} &
        \shortstack{KDA\\heads} &
        \shortstack{Attn. heads\\(Q/KV)} \\
        \midrule
        180M & 181  & 63  & 8  & 768  & 1792 & 6  & 12/3 \\
        250M & 254  & 136 & 16 & 768  & 2048 & 6  & 6/2  \\
        390M & 391  & 234 & 16 & 1024 & 2688 & 8  & 8/2  \\
        510M & 508  & 351 & 24 & 1024 & 2688 & 8  & 8/2  \\
        750M & 755  & 520 & 16 & 1536 & 4096 & 12 & 12/3 \\
        1B   & 1015 & 780 & 24 & 1536 & 4096 & 12 & 12/3 \\
        \bottomrule
    \end{tabular}
\end{table}

We label models by their total parameter counts, including token embeddings,
throughout this appendix: 180M, 250M, 390M, 510M, 750M, and 1B.
Labels are rounded to the nearest 10M. Table~\ref{tab:app-architecture} reports total and nonembedding
parameter counts rounded to the nearest million. Training token-to-parameter
ratios and scaling fits use the unrounded nonembedding parameters.

All models repeatedly stack the four-block sequence
Attention--KDA--KDA--KDA, giving a 3:1 ratio of KDA to softmax-attention
blocks at every model size.

\begin{table}[!h]
    \centering\small
    \renewcommand{\arraystretch}{1.12}
    \caption{Architecture settings shared across KDA language models, in Section~\ref{sec:experimental-details}.}
    \label{tab:app-common-model}
    \begin{tabular}{@{}p{0.29\linewidth}p{0.65\linewidth}@{}}
    \toprule
    Setting & Value \\
    \midrule
    Tokenizer / vocabulary & Qwen3~\citep{yang2025qwen3} / 153,000 entries \\
    Repeating block pattern & Attention--KDA--KDA--KDA; KDA-to-attention ratio 3:1 \\
    Embeddings / FFN & Tied input/output embeddings / SwiGLU~\citep{shazeer2020glu} \\
    Normalization & RMSNorm~\citep{zhang2019rmsnorm}, $\epsilon=10^{-6}$ \\
    KDA head dimension & 128 \\
    KDA value expansion / convolution width & 2 / 4 \\
    KDA output gate & Enabled (sigmoid)~\citep{team2025kimi} \\
    Softmax attention & Grouped-query attention~\citep{ainslie2023gqa}; no projection bias or sliding window \\
    Positional embeddings & Rotary embeddings disabled \\
    \bottomrule
    \end{tabular}
\end{table}

The size-dependent settings are listed in Table~\ref{tab:app-architecture}.

\section{Training Details}
\label{app:scaling_training_details}
\subsection{Data and freezing during early pretraining}
Throughout pretraining, we use DCLM~\citep{li2024datacomplm} with 16,384-token sequences. Texts are tokenized,
terminated with EOS, concatenated, and split into fixed-length sequences;
incomplete remainders of preprocessing batches are discarded.
Table~\ref{tab:app-stages} contrasts the early and remaining portions of pretraining. Vanilla and BLD use the
same data views, token budgets, and optimizer schedules at each model size.

\begin{table}[!h]
    \centering\small
    \renewcommand{\arraystretch}{1.12}
    \caption{Freezing configuration of BLD in pretraining experiments in Section~\ref{sec:experimental-details}.}
    \label{tab:app-stages}
    \begin{tabular}{@{}p{0.23\linewidth}p{0.34\linewidth}p{0.34\linewidth}@{}}
    \toprule
    Setting & Early pretraining & Remaining pretraining \\
    \midrule
    DCLM data view & Global shard 01 & Global shards 02--10 \\
    Token fraction & Approximately $1/11$ (9.09\%) & Approximately $10/11$ (90.91\%) \\
    Vanilla parameters & All trainable & All trainable \\
    BLD: frozen components & Entire KDA blocks: mixer, MLP, and normalization & None \\
    BLD: trainable components & Attention blocks, tied embeddings, and final normalization & All components \\
    Weight / optimizer initialization & Train from scratch & Continue from current weights; reset optimizer and schedule \\
    \bottomrule
    \end{tabular}
\end{table}

Table~\ref{tab:app-budgets} summarizes the token budgets.

\begin{table}[!h]
    \centering\small
    \caption{Training token budgets for the LLM pretraining experiments in Section~\ref{sec:experimental-details}, rounded for readability.}
    \label{tab:app-budgets}
    \begin{tabular}{lrr}
    \toprule
    Model size & Early pretraining tokens & Remaining pretraining tokens \\
    \midrule
    180M & 120M & 1.20B \\
    250M & 280M & 2.80B \\
    390M & 480M & 4.80B \\
    510M & 720M & 7.20B \\
    750M & 1.07B & 10.65B \\
    1B & 1.60B & 16.00B \\
    \bottomrule
    \end{tabular}
\end{table}

Early pretraining uses approximately $2.05N$ tokens, followed by approximately
$20.5N$ tokens in the remaining pretraining, where $N$ is the number of
non-embedding parameters, following the Chinchilla scaling principle~\citep{hoffmann2022training}.

\subsection{Optimization, batching, and restarts}

\begin{table}[!h]
\centering\small
\renewcommand{\arraystretch}{1.12}
\caption{Optimization settings shared by Vanilla and BLD.}
\label{tab:app-optimizers}
\begin{tabular}{@{}p{0.26\linewidth}p{0.3\linewidth}p{0.35\linewidth}@{}}
\toprule
Setting & Early pretraining & Remaining pretraining \\
\midrule
Optimizer & AdamW~\citep{loshchilov2019decoupled} & Muon~\citep{jordan2024muon} with auxiliary AdamW \\
Peak learning rate & $10^{-4}$ & Muon: 0.02; AdamW: 0.001 \\
Minimum learning rate & $10^{-4}$ (no terminal decay) & Muon: 0.001; AdamW: $5\times10^{-5}$ \\
Schedule & Linear warmup, then constant & Linear warmup, constant interval, then cosine decay \\
Adam coefficients & $(0.9,0.95)$ & $(0.9,0.95)$ \\
Muon momentum / iterations & Not applicable & 0.95 / 5 Newton--Schulz iterations \\
Weight decay / clip & 0.1 / global norm 1.0 & 0.1 / global norm 1.0 \\
Precision / seed & BF16 / 0 & BF16 / 0 \\
Global batch (sequences) & 16 & 32 \\
Tokens per update & 262,144 & 524,288 \\
Microbatch per GPU & 1 sequence & 2 sequences \\
\bottomrule
\end{tabular}
\end{table}

\begin{table}[!h]
    \centering\small
    \renewcommand{\arraystretch}{1.12}
    \caption{Size-dependent schedules and batching. Warmup and decay durations are in optimizer updates; accumulation is unchanged throughout pretraining.}
    \label{tab:app-schedules}
    \begin{tabular}{@{}p{0.1\linewidth}p{0.16\linewidth}p{0.16\linewidth}p{0.16\linewidth}p{0.1\linewidth}p{0.12\linewidth}@{}}
    \toprule
    Size & Early warmup & Remaining warmup & Final decay & GPUs & Accum. \\
    \midrule
    180M & 23 & 229 & 229 & 1 & 16 \\
    250M & 53 & 534 & 534 & 1 & 16 \\
    390M & 92 & 916 & 916 & 2 & 8 \\
    510M & 137 & 1,373 & 1,373 & 2 & 8 \\
    750M & 198 & 2,032 & 2,032 & 2 & 8 \\
    1B & 305 & 3,052 & 3,052 & 8 & 2 \\
    \bottomrule
    \end{tabular}
\end{table}

Muon updates eligible hidden matrix weights; embeddings, the output head,
and remaining parameters use auxiliary AdamW. All optimizer groups share
the same multiplicative schedule. Both methods reset optimizer and schedule
state when the early portion ends. 

\section{Evaluation Details}
\label{app:scaling_evaluation_details}
\subsection{Checkpoint and metric selection}
We evaluate the final pretraining checkpoint without fine-tuning, using zero
demonstrations for the downstream language-model evaluation tasks. The decoding strategy for generation tasks is greedy decoding.

\paragraph{Language Modeling Loss}
We report the mean training loss over the final 100 optimizer updates.

\paragraph{Common-Sense Reasoning}
CSR is the macro average of seven tasks: ARC-Easy and ARC-Challenge~\citep{allenai:arc},
HellaSwag~\citep{zellers2019hellaswag}, and PIQA~\citep{Bisk2020} use normalized
accuracy; BoolQ~\citep{clark2019boolq}, WinoGrande~\citep{ai2:winograde}, and LAMBADA~\citep{paperno2016lambada} use accuracy.

\paragraph{Recall}
We use the Based evaluation suite~\citep{arora2024simple} and average the
\texttt{contains} scores over SWDE~\citep{lockard2019openceres},
FDA~\citep{arora2023language}, SQuAD~\citep{rajpurkar2018know},
TriviaQA~\citep{joshi2017triviaqa}, DROP~\citep{dua2019drop}, and
Natural Questions~\citep{kwiatkowski2019natural} with a 2,048-token context.

\paragraph{RULER}
We evaluate RULER~\citep{hsieh2024ruler} at 1K, 2K, 4K, 8K, and 16K tokens, with 100 examples per
task--length combination. These lengths include 10, 11, 12, 12, and 12 tasks,
respectively; QA-1 is included from 2K and QA-2 from 4K. The overall score
averages all 57 task--length combinations.
CSR, Recall, and RULER scores are reported as percentages.

\paragraph{WikiText-2}
We report word-level perplexity on WikiText-2~\citep{merity2016pointer} using the evaluator's
\texttt{word\_perplexity} metric.

\paragraph{LongPPL}
Following LongPPL~\citep{fang2025wrong}, we report pooled perplexity over tokens selected using fixed reference-model
annotations, as detailed below.

\subsection{LongPPL reproduction}

\begin{table}[!h]
    \centering\small
    \renewcommand{\arraystretch}{1.12}
    \caption{Configuration for calculating LongPPL in Section~\ref{sec:experimental-details}.}
    \label{tab:app-longppl-setup}
    \begin{tabular}{@{}p{0.25\linewidth}p{0.68\linewidth}@{}}
    \toprule
    Setting & Value \\
    \midrule
    Dataset / sample & GovReport~\citep{huang2021govreport} / 50 documents \\
    Sampling seed / token limit & 42 / 32,768 per document \\
    Input formatting & Raw text; no chat template or additional BOS \\
    Reference model & Qwen3-8B~\citep{yang2025qwen3}; fixed annotations across compared models \\
    Short context / stride & $K=4096$ / $d=1024$ \\
    Selected targets & 3,920 tokens across 49 documents \\
    Alignment & Same texts and target positions; tokenizer alignment checked \\
    \bottomrule
    \end{tabular}
\end{table}

For token $i$, let $\ell_i^{\rm long}$ and $\ell_i^{\rm short}$ be its reference
log probabilities under the document prefix and a short-context window.
The selected-token indicator is
\begin{equation}
 M_i=\mathbb{1}\{\ell_i^{\rm long}-\ell_i^{\rm short}>2\}
     \mathbb{1}\{\ell_i^{\rm long}>-2\}.
\end{equation}
Target tokens within each stride interval share the same left context boundary, so their context lengths range from $K$ to $K+d-1$ preceding tokens. Table~\ref{tab:app-longppl-setup} lists the fixed sample
and reference settings. The reported metric is
\begin{equation}
 \operatorname{LongPPL}(\theta)=
 \exp\!\left[-\frac{\sum_i M_i\log p_\theta(x_i\mid x_{<i})}
                         {\sum_i M_i}\right].
\end{equation}
We pool selected-token negative log likelihoods before exponentiation; we do
not average document perplexities. The document without selected tokens does
not enter this denominator. 

\section{Detailed Results and Additional Analyses}
\label{app:scaling_results}
\subsection{Downstream task results}
Tables~\ref{tab:app-csr} and~\ref{tab:app-recall} report individual CSR and
Recall scores. The CSR mean here covers the seven tasks listed above, including LAMBADA, and uses the same aggregation as the main tables. Table~\ref{tab:app-ruler} breaks
RULER down by context length; Tables~\ref{tab:app-ruler-tasks-1}--\ref{tab:app-ruler-tasks-3}
provide task-wise averages. All entries use the same final checkpoints as
Tables~\ref{tab:downstream_result} and~\ref{tab:gdn_gla_downstream_result}.

\begin{figure}[!h]
\centering
\includegraphics[width=\linewidth]{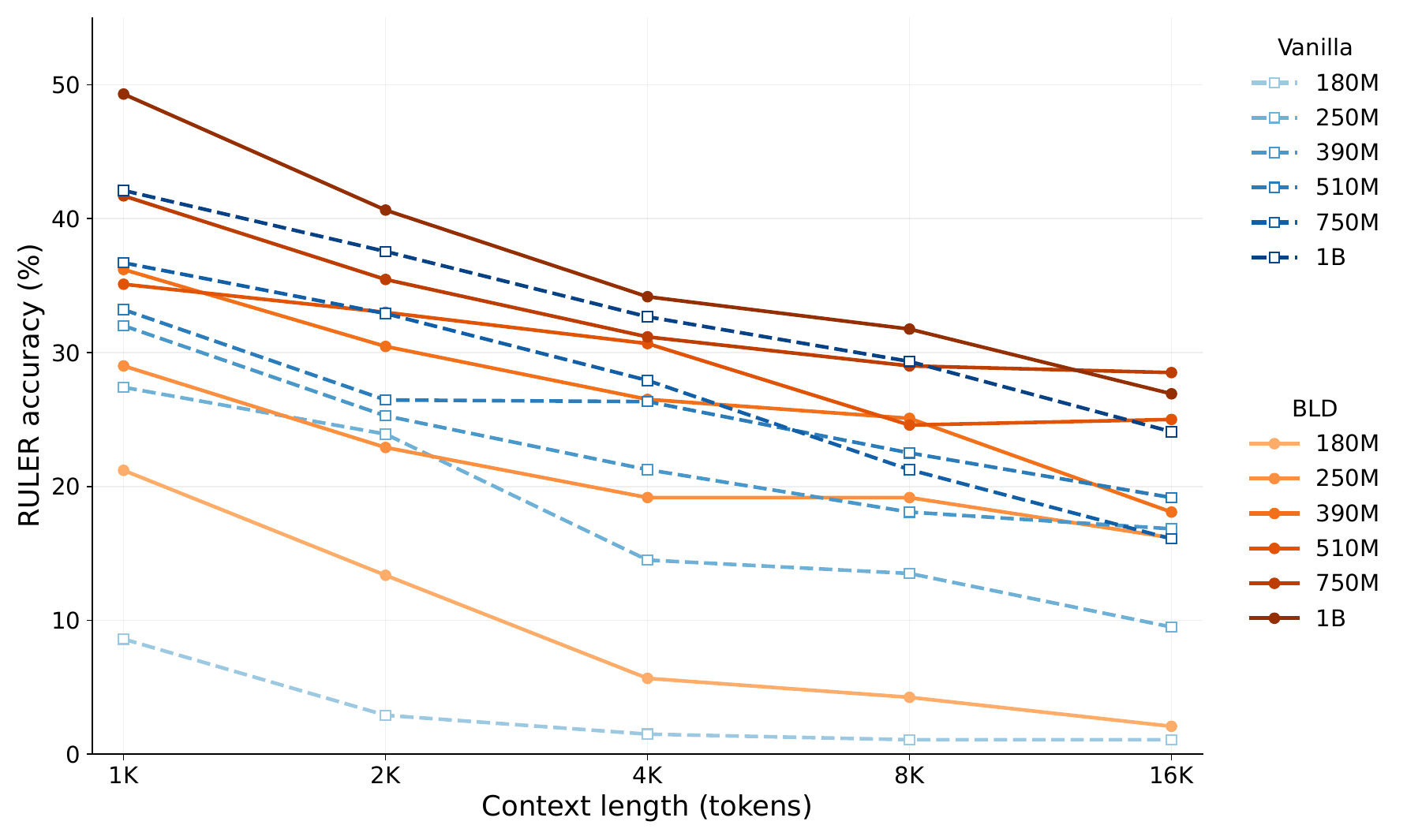}
\caption{RULER accuracy by context length for six KDA model sizes.
Blue dashed lines denote Vanilla; orange solid lines denote BLD. Darker
shades indicate larger models. Each point averages the tasks evaluated at
that length.}
\label{fig:app-ruler-length}
\end{figure}

Figure~\ref{fig:app-ruler-length} compares Vanilla and BLD at each
context length across the six KDA model sizes.

\begin{table}[!h]
    \centering\scriptsize
    \caption{The concrete results of the hybrid models on CSR~(Section~\ref{sec:experimental-details}). LAMBADA reports accuracy. Means use unrounded scores from all seven tasks. GDN and GLA denote the transfer configurations in Appendix~\ref{app:gdn_and_gla}. Unless otherwise specified, all models are KDA-based hybrid models.}
    \label{tab:app-csr}
    \begin{tabular}{llllllllll}
    \toprule
    Size / mixer & Method & ARC-E & ARC-C & Hella. & PIQA & BoolQ & Wino. & LAMBADA & Mean \\
    \midrule
    180M & Vanilla & 36.36 & 22.44 & 27.38 & 58.27 & 58.87 & 51.07 & 9.94 & 37.76 \\
    180M & BLD & 35.65 & 22.44 & 27.64 & 59.30 & 55.84 & 49.96 & 11.59 & 37.49 \\
    \midrule
    250M & Vanilla & 38.43 & 23.98 & 29.80 & 58.98 & 45.05 & 51.38 & 20.57 & 38.31 \\
    250M & BLD & 41.16 & 23.72 & 30.33 & 62.35 & 60.58 & 48.70 & 23.56 & 41.49 \\
    \midrule
    390M & Vanilla & 43.73 & 24.74 & 34.31 & 64.20 & 47.68 & 49.25 & 29.46 & 41.91 \\
    390M & BLD & 43.43 & 24.57 & 35.35 & 65.40 & 58.75 & 50.75 & 31.48 & 44.25 \\
    \midrule
    510M & Vanilla & 46.51 & 26.11 & 36.93 & 66.27 & 58.01 & 50.43 & 29.34 & 44.80 \\
    510M & BLD & 47.31 & 25.17 & 37.42 & 66.97 & 55.69 & 51.22 & 27.61 & 44.49 \\
    \midrule
    750M & Vanilla & 48.06 & 25.51 & 40.68 & 67.19 & 53.76 & 51.14 & 35.26 & 45.94 \\
    750M & BLD & 48.91 & 25.26 & 40.93 & 68.50 & 53.73 & 51.85 & 37.12 & 46.61 \\
    \midrule
    1B & Vanilla & 51.26 & 27.65 & 45.11 & 69.75 & 53.82 & 52.96 & 40.25 & 48.69 \\
    1B & BLD & 51.09 & 27.90 & 45.91 & 69.10 & 55.69 & 55.25 & 41.22 & 49.45 \\
    \midrule
    GDN (400M) & Vanilla & 44.44 & 24.40 & 34.41 & 64.53 & 56.73 & 51.07 & 27.03 & 43.23 \\
    GDN (400M) & BLD & 44.23 & 24.66 & 34.77 & 64.91 & 58.69 & 51.54 & 28.51 & 43.90 \\
    \midrule
    GLA (400M) & Vanilla & 43.43 & 24.06 & 32.43 & 64.25 & 52.23 & 49.80 & 22.10 & 41.19 \\
    GLA (400M) & BLD & 44.53 & 24.74 & 33.32 & 63.44 & 59.48 & 50.36 & 26.57 & 43.20 \\
    \bottomrule
    \end{tabular}
\end{table}

\begin{table}[!h]
    \centering\scriptsize
    \caption{Recall results using the contains metric (percent). Means use unrounded scores. Unless otherwise specified, all models are KDA-based hybrid models.}
    \label{tab:app-recall}
    \begin{tabular}{lllllllll}
    \toprule
    Size / mixer & Method & SWDE & FDA & SQuAD & TriviaQA & DROP & NQ & Mean \\
    \midrule
    180M & Vanilla & 9.09 & 0.54 & 13.40 & 21.74 & 13.08 & 4.50 & 10.39 \\
    180M & BLD & 13.95 & 13.43 & 14.68 & 23.52 & 14.23 & 4.91 & 14.12 \\
    \midrule
    250M & Vanilla & 27.90 & 17.15 & 21.88 & 33.12 & 19.45 & 9.72 & 21.54 \\
    250M & BLD & 29.79 & 30.85 & 25.00 & 35.25 & 18.93 & 12.13 & 25.33 \\
    \midrule
    390M & Vanilla & 34.74 & 26.04 & 27.38 & 38.68 & 20.17 & 13.34 & 26.73 \\
    390M & BLD & 38.34 & 40.83 & 30.09 & 41.35 & 19.84 & 17.14 & 31.27 \\
    \midrule
    510M & Vanilla & 36.72 & 34.21 & 28.92 & 43.36 & 21.99 & 17.77 & 30.50 \\
    510M & BLD & 44.46 & 46.55 & 30.06 & 44.25 & 23.38 & 19.51 & 34.70 \\
    \midrule
    750M & Vanilla & 40.59 & 41.47 & 30.90 & 47.45 & 21.66 & 19.04 & 33.52 \\
    750M & BLD & 49.23 & 50.54 & 33.31 & 49.41 & 21.56 & 21.03 & 37.52 \\
    \midrule
    1B & Vanilla & 44.55 & 42.74 & 33.91 & 52.19 & 24.77 & 20.84 & 36.50 \\
    1B & BLD & 48.15 & 43.83 & 36.76 & 53.08 & 24.20 & 22.81 & 38.14 \\
    \midrule
    GDN & Vanilla & 17.55 & 3.27 & 20.95 & 36.43 & 16.29 & 9.50 & 17.33 \\
    GDN & BLD & 43.74 & 40.74 & 30.06 & 41.88 & 19.50 & 16.34 & 32.05 \\
    \midrule
    GLA & Vanilla & 36.18 & 27.59 & 22.52 & 33.95 & 17.63 & 9.60 & 24.58 \\
    GLA & BLD & 40.86 & 41.20 & 27.78 & 38.15 & 17.97 & 14.73 & 30.12 \\
    \bottomrule
    \end{tabular}
\end{table}

\begin{table}[!h]
    \centering\scriptsize
    \caption{RULER scores by context length (percent). Parentheses give task counts; All averages the 57 task--length entries. Unless otherwise specified, all models are KDA-based hybrid models.}
    \label{tab:app-ruler}
    \begin{tabular}{llllllll}
    \toprule
    Size / mixer & Method & 1K (10) & 2K (11) & 4K (12) & 8K (12) & 16K (12) & All (57) \\
    \midrule
    180M & Vanilla & 8.60 & 2.91 & 1.50 & 1.08 & 1.08 & 2.84 \\
    180M & BLD & 21.20 & 13.36 & 5.67 & 4.25 & 2.08 & 8.82 \\
    \midrule
    250M & Vanilla & 27.40 & 23.91 & 14.50 & 13.50 & 9.50 & 17.32 \\
    250M & BLD & 29.00 & 22.91 & 19.17 & 19.17 & 16.17 & 20.98 \\
    \midrule
    390M & Vanilla & 32.00 & 25.27 & 21.25 & 18.08 & 16.83 & 22.32 \\
    390M & BLD & 36.20 & 30.45 & 26.50 & 25.08 & 18.08 & 26.89 \\
    \midrule
    510M & Vanilla & 33.20 & 26.45 & 26.33 & 22.50 & 19.17 & 25.25 \\
    510M & BLD & 35.10 & 33.00 & 30.67 & 24.58 & 25.00 & 29.42 \\
    \midrule
    750M & Vanilla & 36.70 & 32.91 & 27.92 & 21.25 & 16.08 & 26.53 \\
    750M & BLD & 41.70 & 35.45 & 31.17 & 29.00 & 28.50 & 32.82 \\
    \midrule
    1B & Vanilla & 42.10 & 37.55 & 32.67 & 29.33 & 24.08 & 32.75 \\
    1B & BLD & 49.30 & 40.64 & 34.17 & 31.75 & 26.92 & 36.04 \\
    \midrule
    GDN (400M) & Vanilla & 18.60 & 13.27 & 11.33 & 10.17 & 9.42 & 12.33 \\
    GDN (400M) & BLD & 32.30 & 26.45 & 23.83 & 24.75 & 21.00 & 25.42 \\
    \midrule
    GLA (400M) & Vanilla & 29.00 & 26.55 & 23.42 & 25.08 & 16.92 & 23.98 \\
    GLA (400M) & BLD & 34.30 & 30.64 & 28.58 & 27.58 & 22.75 & 28.54 \\
    \bottomrule
    \end{tabular}
\end{table}

\begin{table}[!h]
    \centering
    \small
    \caption{Task-wise RULER scores averaged over evaluated lengths (percent). V: Vanilla; B: BLD. QA-1 averages four lengths, QA-2 three, and other tasks five.}
    \label{tab:app-ruler-tasks-1}
    \begin{tabular}{l|ll|ll|ll}
    \toprule
    Task & 180M V & 180M B & 250M V & 250M B & 390M V & 390M B \\
    \midrule
    Single-1 & 8.20 & 34.80 & 96.20 & 100.00 & 100.00 & 98.40 \\
    Single-2 & 8.60 & 40.40 & 41.00 & 31.40 & 62.80 & 72.80 \\
    Single-3 & 0.00 & 5.40 & 28.60 & 69.00 & 49.00 & 78.60 \\
    Multi-key-1 & 6.60 & 11.20 & 15.80 & 20.80 & 26.40 & 32.40 \\
    Multi-key-2 & 0.00 & 0.00 & 0.20 & 0.40 & 0.80 & 3.20 \\
    Multi-key-3 & 0.00 & 0.00 & 0.00 & 1.60 & 0.80 & 2.00 \\
    Multi-value & 0.00 & 0.00 & 0.00 & 0.00 & 0.00 & 0.00 \\
    Multi-query & 0.00 & 0.00 & 0.00 & 0.00 & 0.00 & 0.00 \\
    Variable tracking & 0.00 & 0.00 & 0.00 & 0.00 & 0.00 & 0.00 \\
    Frequent words & 0.00 & 0.00 & 0.20 & 0.00 & 0.00 & 3.20 \\
    QA-1 & 4.50 & 4.00 & 9.50 & 11.00 & 10.50 & 9.75 \\
    QA-2 & 9.00 & 9.33 & 13.00 & 12.00 & 10.33 & 13.67 \\
    \bottomrule
    \end{tabular}
\end{table}

\begin{table}[!h]
    \centering
    \small
    \caption{Task-wise RULER scores averaged over evaluated lengths (percent). V: Vanilla; B: BLD. QA-1 averages four lengths, QA-2 three, and other tasks five.}
    \label{tab:app-ruler-tasks-2}
    \begin{tabular}{l|ll|ll|ll}
    \toprule
    Task & 510M V & 510M B & 750M V & 750M B & 1B V & 1B B \\
    \midrule
    Single-1 & 100.00 & 98.80 & 99.60 & 100.00 & 99.80 & 100.00 \\
    Single-2 & 67.80 & 99.00 & 79.60 & 90.40 & 83.00 & 92.80 \\
    Single-3 & 73.80 & 60.20 & 60.60 & 81.40 & 71.60 & 77.00 \\
    Multi-key-1 & 26.60 & 48.40 & 39.20 & 61.00 & 61.80 & 58.40 \\
    Multi-key-2 & 1.00 & 2.80 & 2.80 & 5.40 & 26.60 & 30.80 \\
    Multi-key-3 & 1.60 & 3.80 & 2.20 & 6.60 & 2.60 & 6.60 \\
    Multi-value & 0.00 & 0.00 & 0.00 & 4.20 & 0.40 & 6.80 \\
    Multi-query & 0.00 & 0.00 & 0.00 & 0.40 & 0.40 & 1.20 \\
    Variable tracking & 0.00 & 0.00 & 0.00 & 0.00 & 0.00 & 0.00 \\
    Frequent words & 0.40 & 2.60 & 0.20 & 0.60 & 0.40 & 9.40 \\
    QA-1 & 10.25 & 13.25 & 14.25 & 17.75 & 19.50 & 24.50 \\
    QA-2 & 14.00 & 15.33 & 11.33 & 16.67 & 18.67 & 13.67 \\
    \bottomrule
    \end{tabular}
\end{table}

\begin{table}[!h]
    \centering
    \small
    \caption{Task-wise RULER scores averaged over evaluated lengths (percent). V: Vanilla; B: BLD. QA-1 averages four lengths, QA-2 three, and other tasks five.}
    \label{tab:app-ruler-tasks-3}
    \begin{tabular}{l|ll|ll}
    \toprule
    Task & GDN V & GDN B & GLA V & GLA B \\
    \midrule
    Single-1 & 99.60 & 99.60 & 94.80 & 99.80 \\
    Single-2 & 18.20 & 86.00 & 84.80 & 93.40 \\
    Single-3 & 1.00 & 44.20 & 55.00 & 79.20 \\
    Multi-key-1 & 13.20 & 38.00 & 27.20 & 33.40 \\
    Multi-key-2 & 0.00 & 1.20 & 0.60 & 2.60 \\
    Multi-key-3 & 0.00 & 2.60 & 1.00 & 1.60 \\
    Multi-value & 0.00 & 0.00 & 0.00 & 0.00 \\
    Multi-query & 0.00 & 0.00 & 0.00 & 0.00 \\
    Variable tracking & 0.00 & 0.00 & 0.00 & 0.00 \\
    Frequent words & 0.40 & 0.20 & 0.00 & 0.20 \\
    QA-1 & 4.00 & 11.75 & 4.50 & 7.50 \\
    QA-2 & 8.33 & 14.33 & 10.67 & 15.33 \\
    \bottomrule
    \end{tabular}
\end{table}

\subsection{Scaling fits}
\label{app:scaling_fit}
We use actual nonembedding parameter counts $N$ 
as the model-size variable. A power-plus-constant fit with a shared floor and
method-specific amplitudes and exponents has the form
\begin{equation}
 L_m(N)=E+A_m(N/10^9)^{-\alpha_m},\qquad
 m\in\{\mathrm{Vanilla},\mathrm{BLD}\}.
\end{equation}
The archived fitting procedure minimizes unweighted squared loss residuals
jointly over both methods, with nonnegative floor/amplitudes and positive
exponents. It examines the final-100-update average. 

Table~\ref{tab:app-scaling-fit} reports the fitting using the same
checkpoints as Table~\ref{tab:downstream_result}. Figure~\ref{fig:scaling_laws}
uses the final-100-update fit. Its right panel shows the 1B model's training
loss over 50--100\% of the total pretraining token budget, smoothed
for display by a Gaussian filter with standard deviation 100 updates in the remaining portion
and reflected boundaries.

\begin{table}[!h]
    \centering\small
    \caption{Shared-floor fits for the original-checkpoint six-size series. $A_V,\alpha_V$ are scaling law fitted coefficients for the vanilla model while $A_B,\alpha_B$ are those for the BLD models. RMSE columns are Vanilla/BLD.}
    \label{tab:app-scaling-fit}
    \begin{tabular}{lrrrrrrr}
        \toprule
        Loss window & $E$ & $A_V$ & $\alpha_V$ & $A_B$ & $\alpha_B$ & RMSE~(V) & RMSE~(B) \\
        \midrule
        Final 100 & 2.38867 & 0.45960 & 0.34649 & 0.45266 & 0.34502 & 0.01321 & 0.01424 \\
        \bottomrule
    \end{tabular}
\end{table}

\section{Position-wise Language-Modeling Loss}
\label{app:long_loss}

\begin{table}[!h]
\centering\small
\renewcommand{\arraystretch}{1.12}
\caption{Position-wise language-modeling diagnostic settings.}
\label{tab:app-position-loss-setup}
\begin{tabular}{@{}p{0.23\linewidth}p{0.71\linewidth}@{}}
\toprule
Setting & Value \\
\midrule
Dataset & First 50 nonempty documents from each local Books3 and BookCorpus2 cache from The Pile~\citep{gao2020pile} (100 total) \\
Input processing & Independent documents; no added special tokens; cached token IDs and checked tokenization \\
Input / prediction limit & 16,384 input tokens / 16,384 predictions per document \\
Execution & Full decoder prefix; chunked output projection; original position settings \\
Precision & BF16 model execution; FP32 logits \\
\bottomrule
\end{tabular}
\end{table}

\begin{figure}[!h]
    \centering
    \includegraphics[width=1\linewidth]{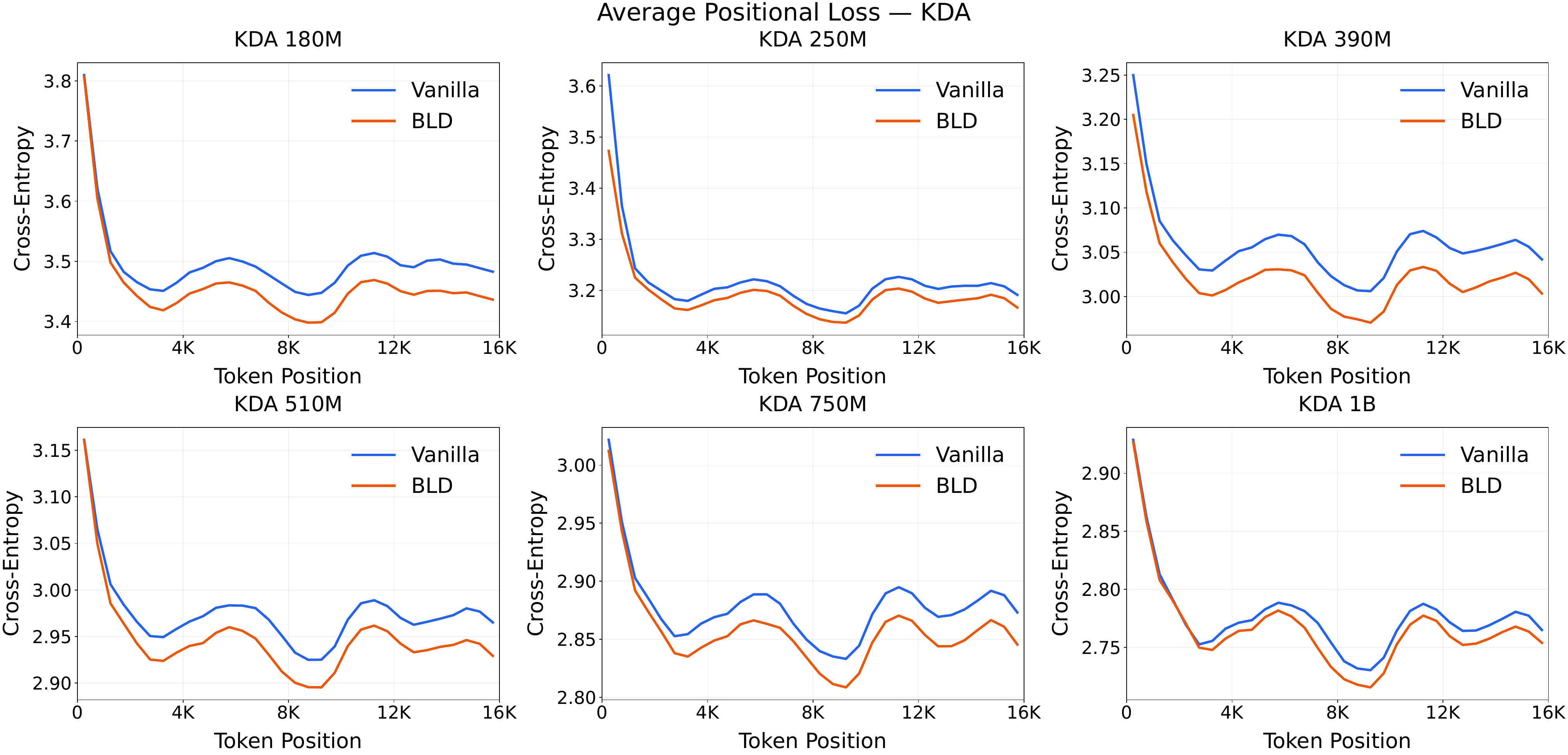}
    \caption{Average loss of KDA hybrid models of different scales, as functions of token positions.}
    \label{fig:all_long_loss}
\end{figure}

Table~\ref{tab:app-position-loss-setup} describes the diagnostic underlying
Figure~\ref{fig:long_loss}. Figure~\ref{fig:all_long_loss} shows the results of the same diagnostic across all model sizes.

\section{Transfer to GDN and GLA}
\label{app:gdn_and_gla}

\begin{table}[!h]
\centering\small
\renewcommand{\arraystretch}{1.12}
\caption{Configurations for the GDN and GLA models in Section~\ref{sec:transfer-to-other-architectures}. Parameter and token counts are rounded; M and B denote million and billion.}
\label{tab:app-transfer-config}
\begin{tabular}{@{}p{0.33\linewidth}p{0.28\linewidth}p{0.28\linewidth}@{}}
\toprule
Setting & GDN & GLA \\
\midrule
Blocks / hidden / FFN width & 16 / 1,024 / 2,688 & 16 / 1,024 / 2,688 \\
Attention block indices & 0, 4, 8, 12 & 0, 4, 8, 12 \\
Attention Q / KV heads & 8 / 2 & 8 / 2 \\
Recurrent blocks / heads & 12 / 8 & 12 / 8 \\
Recurrent dimensions & Head dimension 128 & Key expansion 1 \\
Value expansion & 2 & 2 \\
Convolution width / output gate & 4 / enabled & 4 / enabled \\
Rotary embeddings & Disabled & Disabled \\
Total parameters & 400M & 401M \\
Nonembedding parameters & 244M & 244M \\
Early / remaining updates & 1,831 / 9,155 & 1,831 / 9,155 \\
Early pretraining tokens & 480M & 480M \\
Remaining pretraining tokens & 4.80B & 4.80B \\
\bottomrule
\end{tabular}
\end{table}

Table~\ref{tab:app-transfer-config} reports the GDN~\citep{yang2025gated} and GLA~\citep{yang2023gated} replacement architectures.
The 400M designation in Table~\ref{tab:gdn_gla_downstream_result} identifies
the reference KDA configuration. They match KDA in depth, width, and token budget. Optimization and full-block freezing/unfreezing
follow the 390M KDA configuration. Data budgets remain fixed rather than
being rescaled to each replacement's parameter count. Detailed downstream scores appear in Tables~\ref{tab:app-csr}--\ref{tab:app-ruler}.

\section{Retrieval Head Identification and Retrieval Evaluation}
\label{app:alpha-value}

\paragraph{Retrieval Head identification.}
We apply a DuoAttention-style retrieval head identification procedure~\citep{xiao2025duoattention} to the final 510M
Vanilla and BLD checkpoints. Each model has six attention layers, with two
KV heads and eight query heads per layer. We learn one gate
$\alpha_{\ell,g}\in[0,1]$ per KV head (12 gates in total), shared by its four
query heads. Before the attention output projection, each head interpolates
between full causal and streaming attention:
\begin{equation}
  \widetilde{\mathbf{o}}_{\ell,h}
  = \alpha_{\ell,g(h)}\mathbf{o}^{\mathrm{full}}_{\ell,h}
  + (1-\alpha_{\ell,g(h)})\mathbf{o}^{\mathrm{stream}}_{\ell,h}.
\end{equation}
Streaming attention retains the first 128 tokens and the most recent
256 tokens, subject to causality. All pretrained parameters are frozen;
KDA layers retain their original computation without gates.

Each model serves as its own full-attention teacher. We optimize the gates
using mean squared error between teacher and student final hidden states
at reference-answer token positions, plus
$0.05\,\mathrm{mean}(|\alpha|)$. Both branches receive reference answers
during identification; this is hidden-state distillation, not retrieval
fine-tuning of the base model. Gates start at one and are clamped to
$[0,1]$ after each update. Table~\ref{tab:alpha-settings} summarizes the
main settings.

\begin{table}[!h]
  \centering
  \small
  \caption{Settings for retrieval head identification experiments in Section~\ref{sec:retrieval-head}.}
  \label{tab:alpha-settings}
  \begin{tabular}{ll}
    \toprule
    Setting & Value \\
    \midrule
    Optimization steps / batch size & 1,000 / 1 sequence \\
    Optimizer & AdamW, no weight decay \\
    Peak / minimum learning rate & 0.02 / 0.002 \\
    Schedule & 200-step warmup and final 200-step decay \\
    Sequence length & 16,384 tokens \\
    Passkeys per sequence & 10, each containing 32 words \\
    Background / insertion depth & BookSum / 5--95\% \\
    \bottomrule
  \end{tabular}
\end{table}

\paragraph{Masking and evaluation.}
Within each model, we rank all 12 KV heads by increasing learned alpha
and progressively replace full attention with streaming attention,
starting with the lowest-alpha heads. Remaining heads use full attention;
the continuous alpha values are used only for ranking. No further training
is performed. Thus, a ``masked head'' denotes a KV head whose four query
heads use restricted context, rather than a head with zeroed output.
The evaluated budgets are $0,1,2,4,5,6,7,8,10,11,12$ KV heads, obtained by
rounding masking ratios of $0\%,10\%,\ldots,100\%$.

We use the same separately generated set of 64 documents and 10 passkeys
per document for both models and every budget (640 queries per condition).
Each passkey contains 32 random NATO-alphabet words embedded in BookSum
text. The 16K length includes the document and reference question--answer
pairs. At evaluation, each question is answered independently from the
document cache using greedy decoding, without reference answers or earlier
question--answer pairs. The plotted accuracy is the percentage of queries
whose \emph{first generated token} matches the reference.

\subsection{Details of the Gradient Dynamics Experiment}
\label{app:grad-change}

Figure~\ref{fig:grad_change} compares a never-frozen
baseline with a KDA-freezing intervention under the
settings in Table~\ref{tab:gradient_settings}.
At the branching point, the frozen trajectory is split
into continued-freezing and unfreezing branches with
identical model, optimizer, scheduler.
Unfreezing resumes the paused KDA Adam states without
resetting the optimizer. Within each seed, all compared
branches follow the same training-data sequence.
Training uses full backward passes and the same
step-specific baseline clipping coefficient before
frozen updates are masked.

We measure unclipped attention gradients using the same
gradient-enabled diagnostic graph across conditions.
For each attention mixer, we average the per-example
gradient vectors before taking their $\ell_2$ norm,
then divide by the corresponding baseline norm at the
same step. The plotted statistic is the arithmetic
mean of the two attention mixers' trajectories, each
computed as the geometric mean of paired ratios across
seeds. The left panel enlarges the freezing phase of
the full trajectory shown on the right.

\begin{table}[!h]
    \centering
    \small
    \caption{Experimental settings for
    Figure~\ref{fig:grad_change}.}
    \label{tab:gradient_settings}
    \begin{tabular}{ll}
    \toprule
    Setting & Value \\
    \midrule
    Mixer order & KDA--Attention--KDA--Attention \\
    Total model parameters & 3,555,588 \\
    Independent seeds & 10 \\
    Frozen modules & Both KDA mixers \\
    Frozen parameter count & 1,453,060 \\
    Measured parameters & 262,144 per attention mixer \\
    Freezing begins & After update 175 \\
    Frozen/unfrozen branches split & After update 225 \\
    Final measured step & 425 \\
    Gradient probe & 32 fixed training examples \\
    Retrieval gap & 8192 \\
    Probe loss & First-answer-token cross-entropy \\
    Measurement frequency & Every training step \\
    Temporal smoothing & None \\
    \bottomrule
    \end{tabular}
\end{table}

\end{document}